\pdfoutput=1
\documentclass[onecolumn]{fairmeta}

\ifdefined\pdfminorversion
\fi

\usepackage{booktabs}
\usepackage{amsmath}
\usepackage{amssymb}
\usepackage{graphicx}
\usepackage{array}
\usepackage{tabularx}
\usepackage{colortbl}
\usepackage{multirow}
\usepackage{enumitem}
\usepackage{placeins}
\usepackage{float}
\usepackage{hyperref}

\newcommand{\method}{FreeSpeed}

\newcommand{\methodstack}{\shortstack{Free\\[-2pt]Speed}}
\newcommand{\vanillastack}{vanilla}
\newcommand{\wocrstack}{\shortstack{w/o\\[-2pt]CR}}
\definecolor{SimLavender}{RGB}{241,247,252}
\definecolor{RRCBlue}{RGB}{224,237,248}
\definecolor{GroupBand}{RGB}{245,239,251}
\newcolumntype{Y}{>{\centering\arraybackslash}X}
\newcolumntype{Z}{>{\columncolor{RRCBlue}\centering\arraybackslash}X}


\title{FreeSpeed: Training-Free Speed Control for Generative Robot Policies}
\author[1,*]{Yuxuan Hu}
\author[1,*]{Shilin Shan}
\author[1]{Qiheng Wang}
\author[1]{Jinghan Yang}
\author[1]{Junqiao Fan}
\author[2]{Hao Wan}
\author[1,\dagger]{Jianfei Yang}

\renewcommand\affiliation[2][]{\addtolist[#1]{#2}{\affiliationlist}{\affiliationformat}{\qquad}}
\affiliation[1]{MARS Lab, Nanyang Technological University}
\affiliation[2]{ROKAE Robotics}
\contribution[*]{Equal Contribution}
\contribution[\dagger]{Corresponding Authors}

\correspondence{Jianfei Yang at \email{jianfei.yang@ntu.edu.sg}.}
\metadata[Project Page]{\url{https://yuxuanhu9.github.io/FreeSpeed/}}

\hypersetup{
  pdftitle={FreeSpeed: Training-Free Speed Control for Generative Robot Policies},
  pdfauthor={Yuxuan Hu, Shilin Shan, Qiheng Wang, Jinghan Yang, Junqiao Fan, Hao Wan, Jianfei Yang},
  pdfsubject={Training-free speed control for generative robot policies},
  pdfkeywords={Robot Learning, Imitation Learning, Generative Policies, Speed Control, Training-Free Adaptation}
}

\abstract{
Online control of execution speed is essential for deploying robot policies in real-world scenarios, as robots may need to speed up under time constraints or slow down to facilitate human interaction and improve safety. However, imitation-learned policies inherit the execution speed of their demonstrations, and test-time speed modification can introduce unrecoverable out-of-distribution observations, reducing task success. We observe that the directional inconsistency of action chunks reflects task-phase criticality, indicating how aggressively action step lengths can be modified while preserving task success. Based on this observation, we introduce \textbf{FreeSpeed}, a training-free module that post-processes action chunks from pretrained policies. FreeSpeed resamples each predicted chunk at the requested rate, then uses directional inconsistency between adjacent actions as the primary signal for rescaling. This signal adaptively determines how closely the execution speed can approach the requested speed, allowing flexible speed adjustment within the evaluated limits without compromising task success. Across three policy families and 50 simulated tasks, \method{} supports online speed changes, with realized execution rates spanning \(0.22\times\) to \(2.53\times\) among settings that preserve per-task success. Across four real-world manipulation tasks, \method{} achieves an average success rate of \(94.0\)\%, matching the frozen policy's \(93.8\)\%, while realizing execution rates from \(0.38\times\) to \(1.97\times\).
}

\begin{document}

\maketitle

\begin{figure}[H]
  \centering
  \includegraphics[width=0.97\textwidth]{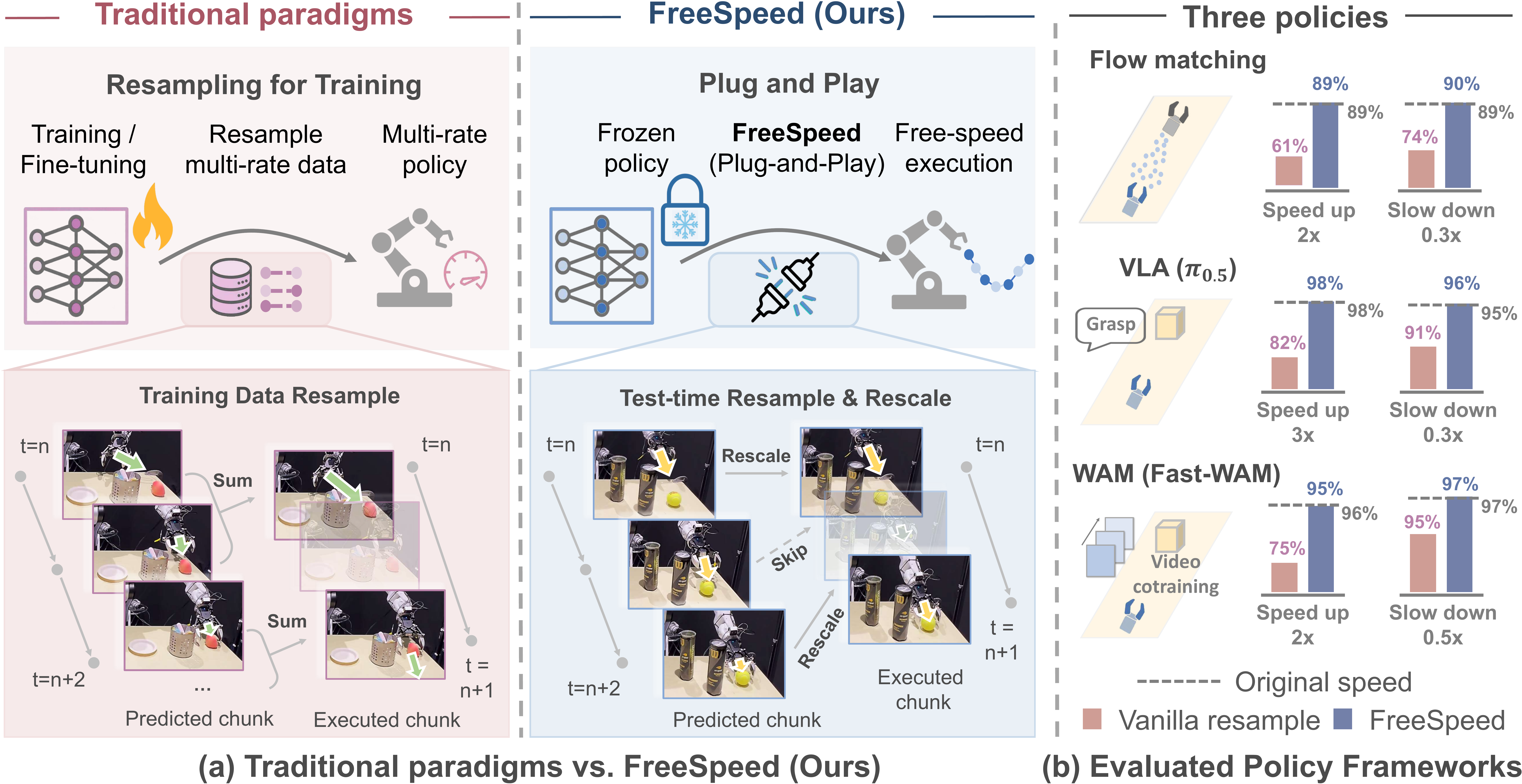}
  \caption{\textbf{FreeSpeed modulates the execution speed of a frozen policy.} (a) Methods based on demonstration retiming resample demonstrations and train or fine-tune the policy, while \method{} resamples and rescales predicted action chunks at test time. (b) Success rates of \method{} and vanilla resampling on three policies under speedup and slowdown commands.}

  \label{fig:teaser}
\end{figure}

\section{Introduction}

\label{sec:introduction}

Modulating execution speed online matters when robots operate across changing
tasks and deployment conditions.
Fast motion improves efficiency in repetitive manipulation \citep{tossingbot}, whereas slow motion can be essential for safe and comfortable human--robot interaction in shared workspaces \citep{lachner2021,adaptive_hri}, particularly as robots are increasingly deployed in diverse human-centered tasks \citep{fan2026mmhri,lu2026fall,hu2026activevital,hu2026waveman}. Within a single manipulation task, a robot may need to move quickly during routine phases, such as object transport, and slow down during phases that require precision, such as object placement and grasping \citep{demospeedup,espada,speedtuning}. A deployed policy should therefore accept speed requests during execution rather than operate at a single pace fixed during training.

Imitation-based policies learn the execution speed implicitly from demonstrations \citep{shi2026diversity}, and they do not normally expose the speed control interface. Despite the impressive speed-modification performance of existing methods \citep{speedtuning,speedaug,speedup_patch,espada,autospeed,tempovla,demospeedup}, almost all require a training process closely tied to task-specific data augmentation. This process typically involves specifying speed factors, post-processing collected demonstrations, and then fine-tuning or retraining a policy in a speed-informed manner. This dependence can limit the methods' adaptability to unseen scenarios, making speed control feasible but difficult to generalize.

Therefore, it is necessary to explore training-free, model-independent speed control that can transfer directly across tasks and pretrained models. A straightforward approach is to downsample or upsample the predicted action chunk at test time. At a fixed command frequency, combining or skipping action steps accelerates execution by reducing the number of executable steps, while interpolation slows execution by adding intermediate steps. Despite its simplicity, this approach changes step lengths, which can induce out-of-distribution (OOD) observations \citep{yang2024generalized, sail, race, speedaug} and cause subsequent trajectory predictions to drift \citep{pmlr-v15-ross11a} (Section~\ref{sec:in_distribution}). This is particularly risky during task phases that require precision, such as grasping and placing, where execution errors at critical steps can cause the entire task to fail.

To this end, we identify directional inconsistency as a test-time indicator of task-phase criticality. In the studied tasks, inconsistency tends to be high during critical phases and low otherwise. Adjusting action step lengths according to this signal helps preserve task success while allowing more aggressive speed changes during less critical phases. Building on this insight, we introduce \method{}, a training-free, plug-and-play module for frozen action-chunking policies. \method{} also builds on action resampling but adapts the command step length according to the directional consistency of the predicted action chunk. It pushes execution speed toward its limit when consecutive action directions remain consistent and gradually shifts execution back toward the original speed as inconsistency increases. It uses only the policy's predicted motion and requires no policy updates, additional data, or a learned scheduler. The same interface accepts both fixed and time-varying rate commands. The contributions of this work are as follows.

\begin{itemize}[leftmargin=1.25em]

\item We use adjacent-action cosine inconsistency as a training-free proxy for phase criticality and show in Section~\ref{sec:in_distribution} that adapting step-length rescaling to this signal keeps the frozen policy's observations close to the training distribution under both acceleration and deceleration.

\item We introduce \method{}, a training-free module that combines temporal resampling with continuous inconsistency-dependent action scaling for fixed and time-varying speed control.

\item We evaluate \method{} on the frozen $\pi_{0.5}$, Fast-WAM, and task-specific Flow Matching policies, across 50 LIBERO tasks and four real-robot tasks, under fixed and time-varying commands from $0.2\times$ to $4\times$. In simulation, among task-rate settings that preserve each frozen policy's per-task success, realized rates span $0.22\times$ to $2.53\times$. Across four real-robot tasks, it achieves an average success rate of $94.0$\%, compared with the frozen policy's $93.8$\%, with realized execution rates ranging from $0.38\times$ to $1.97\times$.

\end{itemize}

\section{Related Work}
\label{sec:related_work}

Existing methods differ in which components are modified to enable speed control. They can be broadly grouped into three approaches: (1) incorporating speed variation into policy learning; (2) adapting both policy learning and the execution framework; and (3) training auxiliary speed-control modules while keeping the base policy frozen.

The first group modifies the policy through demonstration retiming \citep{demospeedup, espada, tempovla}, rate-aware objectives or architectures \citep{autospeed, hipolicy}, or speed-augmented pretraining followed by RL fine-tuning \citep{speedaug}. These methods primarily target faster execution, with limited attention to deceleration or user control. TempoVLA \citep{tempovla} supports user-specified rates but requires training to enable this capability.

The second group combines policy adaptation with precise tracking. SAIL and RACE process collected demonstrations, use reached states as action prediction targets, and combine execution retiming with high-gain control \citep{sail, race}. SAIL's ablation indicates that high-gain control contributes substantially to its acceleration gains. Building on this approach, \citet{realtime_vla_v2} further learn faster-than-demonstration execution from human-modulated rollouts.

The third group freezes the base policy and trains an external speed controller. SpeedTuning trains a speed policy online via reinforcement learning \citep{speedtuning}, whereas SuP trains a downsampling scheduler offline using a learned world model \citep{speedup_patch}. Both require additional training and select execution rates autonomously rather than follow user-specified rates.

In contrast, \method{} adjusts a frozen policy's output toward rates specified by the user in real time, supporting both acceleration and deceleration. It requires no post-processing of training data, no speed-informed retraining, and no change to the low-level control law or controller gains. Table~\ref{tab:capabilities} summarizes these capabilities and compares \method{} with the aforementioned methods.

\begin{table}[!t]
\centering
\setlength{\belowcaptionskip}{4pt}
\caption{\textbf{Speed-control capabilities of the compared methods.} A filled dot indicates that the capability is demonstrated in the corresponding paper; an open dot indicates that it is not demonstrated.}
\label{tab:capabilities}
\small
\setlength{\tabcolsep}{4pt}
\renewcommand{\arraystretch}{0.95}
\newcommand{\yes}{$\bullet$}
\newcommand{\no}{\textcolor{black!28}{$\circ$}}
\begin{tabularx}{\textwidth}{@{}lYYYYYY@{}}
\toprule
 & No rate & Unmodified & Unchanged & User-set & Slow- & In-episode \\
 & training & demos & controller & rate & motion & rate \\
\midrule
SuP \citep{speedup_patch}       & \no  & \no  & \no  & \no  & \no  & \no  \\
SpeedTuning \citep{speedtuning} & \no  & \yes & \yes & \no  & \no  & \no  \\
TempoVLA \citep{tempovla}       & \no  & \no  & \yes & \yes & \yes & \yes \\
SAIL \citep{sail}               & \no  & \no  & \no  & \no  & \no  & \no  \\
RACE \citep{race}               & \yes & \no  & \no  & \no  & \no  & \no  \\
DemoSpeedup \citep{demospeedup} & \no  & \no  & \no  & \no  & \no  & \no  \\
AutoSpeed \citep{autospeed}     & \no  & \yes & \no  & \no  & \yes & \no  \\
SpeedAug \citep{speedaug}       & \no  & \no  & \yes & \no  & \no  & \no  \\
\rowcolor{RRCBlue}
\method{} (ours)                & \yes & \yes & \yes & \yes & \yes & \yes \\
\bottomrule
\end{tabularx}
\end{table}

\section{Problem Formulation}
\label{sec:problem_formulation}

An action-chunking policy $\pi_\theta$, parameterized by $\theta$, takes the current observation $o_t$ at replanning event $t$ and an optional language instruction $\ell$ as input and predicts an action chunk. We denote the policy-specific execution prefix, consisting of its first $H$ actions, as
\begin{equation}
  A_t = \pi_{\theta}(o_t,\ell)
      = \left(a_t^1,\ldots,a_t^H\right),
  \qquad
  a_t^i = \left[\Delta p_t^i,\Delta r_t^i,g_t^i\right],
  \label{eq:action_chunk}
\end{equation}
where $i\in\{1,\ldots,H\}$ indexes actions within the execution prefix. $\Delta p_t^i\in\mathbb{R}^{3}$ and $\Delta r_t^i\in\mathbb{R}^{3}$ are translation and rotation increments, respectively, and $g_t^i$ is the gripper command. Under the original execution rule, these actions are queued and executed as robot commands at a fixed control interval $\Delta t$ between successive commands \citep{act,diffusion_policy,team2024octo}. 

We seek to control the execution rate through two mechanisms. First, skipping or repeating predicted actions changes the time required to execute a chunk at a fixed robot control frequency. Second, scaling the translation and rotation increments changes the trajectory length covered by each chunk. Both the resampling and rescaling effects should be adjustable through user input between successive policy inferences to accommodate real-time speed commands.

The user specifies the desired execution rate through a command $\rho_t>0$, where $\rho_t=1$ denotes the original rate, $\rho_t>1$ requests faster execution, and $\rho_t<1$ requests slower execution. The same value acts as the stride at which the predicted chunk is resampled. The performance of execution rules is evaluated using task success rate and realized execution rate $\widehat{r}$, defined as the ratio of the steps used to complete a task under the policy's original execution to those used after applying the rule. The objective is for $\widehat{r}$ to approach $\rho_t$ while preserving the policy's original success rate.

\section{\method{}: Inconsistency-Gated Chunk Post-Processing}
\label{sec:method}

\begin{figure}[t]
  \centering
  \includegraphics[width=0.98\textwidth]{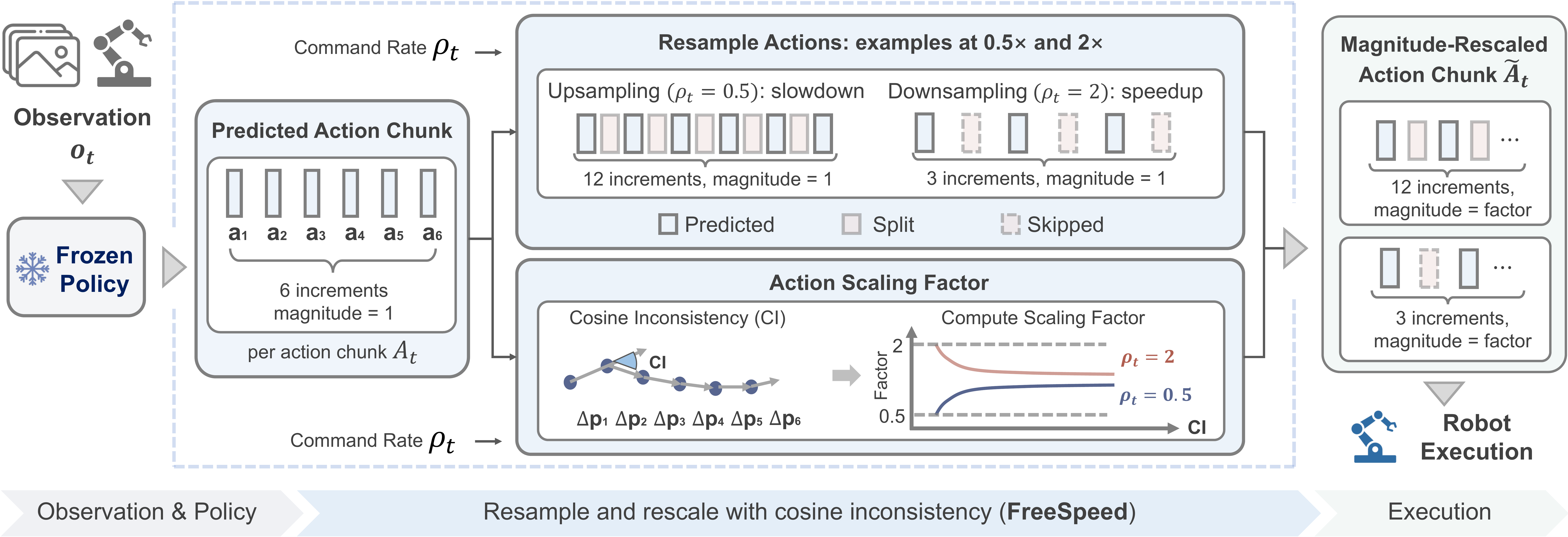}

  \caption{\textbf{FreeSpeed at one replan.} The frozen policy predicts an action chunk. Its translation increments provide a cosine inconsistency score, and the rate command determines temporal resampling. Together, the score and command determine the factor used to rescale the resampled translation and rotation increments.}

  \label{fig:method}
\end{figure}

At each replan, \method{} resamples the predicted chunk of incremental (delta) actions on a temporal grid with stride $\rho_t$ and scales the resampled actions by a factor $f_t$. Under this scheme, acceleration skips predicted actions, while deceleration splits each action. The remainder of this section formalizes these operations and introduces the mapping from $\rho_t$ to $f_t$. Figure~\ref{fig:method} visualizes the execution loop.

\subsection{Why Cosine Inconsistency and Its Formulation}

Naively modifying the execution rate of predicted actions alters the distribution of the resulting observations. Intuitively, speeding up makes observations sparser along the executed trajectory, whereas slowing down makes them denser. This sometimes shifts the policy's inputs away from the training distribution, leading to action-chunk predictions that are poorly supported by the training data. We further investigate and quantify this effect in Section~\ref{sec:in_distribution}. The results suggest that the shift in observation distributions can in turn affect action predictions and, consequently, task success.

We further observe that chunk geometry is a determining factor in whether the observation distribution shifts under speed modification. Along straight segments, the original and modified chunks produce similar observation distributions. However, when a chunk contains turns or noise, these distributions differ substantially, as further illustrated in Figure~\ref{fig:ood_grid}c. Therefore, to maintain the original success rate, speed adjustment must account for the consistency of its action directions.

For consecutive translation increments, we define cosine inconsistency as:
\begin{equation}
  d_{\cos,t}^i
  = 1-
    \frac{\left\langle\Delta p_t^i,\Delta p_t^{i+1}\right\rangle}
         {\left\lVert\Delta p_t^i\right\rVert_2
          \left\lVert\Delta p_t^{i+1}\right\rVert_2},
  \label{eq:pair_cosine_inconsistency}
\end{equation}
and aggregate the chunk with
\begin{equation}
  I_{\cos}(A_t)=\max_{i} d_{\cos,t}^i.
  \label{eq:chunk_cosine_inconsistency}
\end{equation}
Here, $\langle\cdot,\cdot\rangle$ denotes the vector dot product, and $d_{\cos,t}^i$ measures the change in direction between increments $i$ and $i+1$. As defined in Equation~(3), the cosine inconsistency of an action chunk is the maximum value of $d_{\cos,t}^i$ over its $H-1$ adjacent increment pairs.

\subsection{Resampling and Rescaling with Cosine Inconsistency}

As discussed above, at a fixed robot control frequency, execution speed can be modified by resampling action chunks represented as incremental (delta) actions. For acceleration, some predicted actions can be skipped with the retained actions rescaled, reducing the number of command steps needed to cover the distance traveled by the original chunk. Alternatively, multiple actions can be summed into a larger step to achieve the same goal \citep{speedup_patch}. For deceleration, each original action can be split into multiple sub-actions that preserve its direction while reducing the step length. We compare against these naive speed-modification baselines in Table~\ref{tab:sim_fixed}.

The resampling process is formalized as:
\begin{equation}
  \overline{A}_t=\mathcal{R}_{\rho_t}(A_t)
  =\left(\overline{a}_t^1,\ldots,\overline{a}_t^{M_t}\right),
  \label{eq:temporal_resampling}
\end{equation}
where $\mathcal{R}_{\rho_t}$ denotes temporal resampling of the predicted chunk on a grid of stride $\rho_t$. Here $\rho_t$ is the commanded stride and $M_t$ is $H/\rho_t$ rounded to the nearest integer, and $m\in\{1,\ldots,M_t\}$ indexes actions within the resampled chunk. However, as discussed in Section~4.1, resampling can shift the observation distribution, making subsequently predicted actions less controllable. Such shifts can cause failure during critical task phases, particularly when grasping or releasing an object.

To this end, we use cosine inconsistency (CI) to guide rescaling and mitigate the observation distribution shifts caused by naive speed modification. Intuitively, when CI is small, action steps are scaled toward the commanded speedup or slowdown limit, as the resulting distribution shift is less likely to disrupt execution. When CI is large, more conservative scaling is applied, as this may indicate a critical task phase in which distribution shifts can cause failure.

Formally, for each commanded rate $\rho_t$, we fix two scaling parameters, $f_{\mathrm{s}}(\rho_t)$ and $f_{\mathrm{c}}(\rho_t)$, which together define the bounds on the rescaling factor. Their values are listed in Appendix~\ref{app:implementation}. For acceleration ($\rho_t>1$), $f_{\mathrm{s}}>f_{\mathrm{c}}\geq1$. For deceleration ($\rho_t<1$), $f_{\mathrm{s}}=\rho_t<f_{\mathrm{c}}\leq1$. During execution, the factor $f_t$ equals $f_{\mathrm{s}}$ when $I_{\cos}$ is zero and approaches $f_{\mathrm{c}}$ as $I_{\cos}$ increases. Action steps in chunks with large $I_{\cos}$ are therefore scaled more conservatively, with a factor closer to $f_{\mathrm{c}}$, whereas those in chunks with small $I_{\cos}$ are scaled more aggressively, with a factor closer to $f_{\mathrm{s}}$:
\begin{equation}
  f_t
  = F\!\left(\rho_t,I_{\cos}(A_t)\right)
  = f_{\mathrm{c}}(\rho_t)
    + \bigl(f_{\mathrm{s}}(\rho_t)-f_{\mathrm{c}}(\rho_t)\bigr)
    \exp\!\left(-\lambda I_{\cos}(A_t)\right),
  \label{eq:factor_mapping}
\end{equation}
where $\lambda>0$ determines how quickly $f_t$ moves from $f_{\mathrm{s}}(\rho_t)$ toward $f_{\mathrm{c}}(\rho_t)$ and is shared across all stride settings. The sensitivity analysis in Figure~\ref{fig:sensitivity} evaluates $\lambda$ and $f_{\mathrm{s}}$, showing a broad range of effective settings for both.

For $\overline{a}_t^m=[\overline{\Delta p}_t^m, \overline{\Delta r}_t^m,\overline{g}_t^m]$, \method{} executes
\begin{equation}
  \widetilde{a}_t^m
  = \left[
      f_t\overline{\Delta p}_t^m,
      f_t\overline{\Delta r}_t^m,
      \overline{g}_t^m
    \right].
  \label{eq:rrc_action}
\end{equation}
The frozen policy then replans from the next observation. Recomputing $I_{\cos}(A_t)$ and $f_t$ at every replan allows updated stride commands to take effect in the next inference cycle.

\section{Experiments}

\label{sec:experiments}

\begin{figure}[t]
  \centering
  \includegraphics[width=0.97\textwidth]{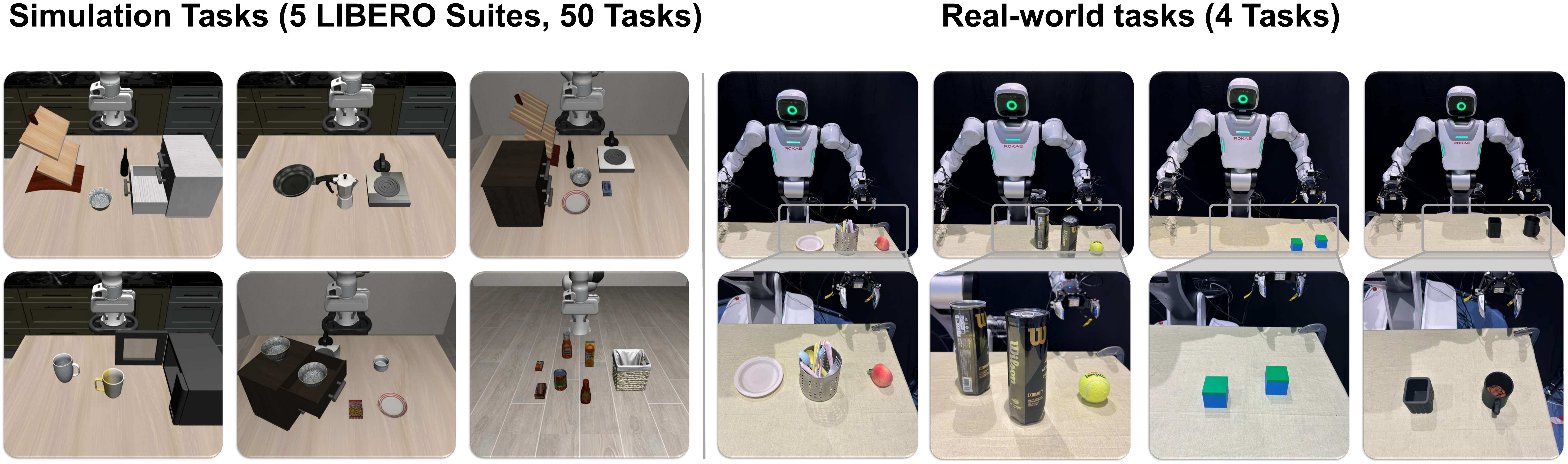}

  \caption{\textbf{Evaluation environments.} 50 simulation tasks from LIBERO and four real-robot tasks.}
  \label{fig:evaluation_environments}

\end{figure}

\subsection{Evaluation Protocol}
\label{sec:evaluation_protocol}

We evaluate three policy architectures: VLA, WAM, and task-specific flow matching. We use checkpoints of $\pi_{0.5}$ \citep{pi05,pi0} and Fast-WAM \citep{fastwam} that were pretrained on 40 tasks spanning LIBERO-Spatial, LIBERO-Object, LIBERO-Goal, and LIBERO-Long \citep{libero}. We keep these policies frozen and evaluate FreeSpeed using their action predictions. In addition to these released checkpoints, we train task-specific flow-matching policies \citep{flow_matching} on 10 LIBERO-90 tasks and 4 real-robot tasks. Figure~\ref{fig:evaluation_environments} shows representative simulation and real-robot task setups.

\subsection{Cosine Inconsistency Tracks the Task Phase}
\label{sec:phase}

\begin{figure}[t]
  \centering
  \includegraphics[width=0.90\textwidth]{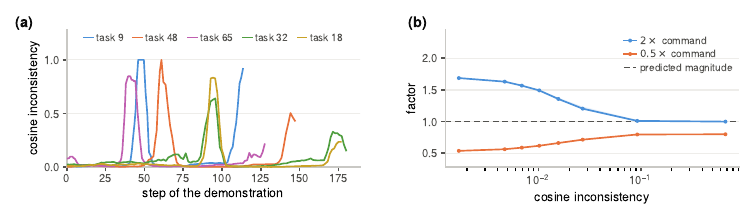}

  \caption{\textbf{The signal and the response.} (a) Cosine inconsistency along human demonstrations of five LIBERO-90 tasks, with peaks during grasping and placement. (b) The mapping from cosine inconsistency to the rescaling factor under the $2\times$ and $0.5\times$ commands.}

  \label{fig:demo_phase}
\end{figure}

We illustrate how $I_{\cos}$ varies with task progress using five LIBERO-90 tasks as examples, each with roughly $50$ demonstrations. For every demonstration, we slide a window of length $H$ along the trajectory and compute $I_{\cos}$ within each window. Figure~\ref{fig:demo_phase}a plots one curve per task. 

Segments with high $I_{\cos}$ correspond to decisive task phases, identified in these examples as grasping and placement in pick-and-place tasks. This partly reflects how human operators behave during data collection: they slow down and hesitate during critical phases and move faster otherwise \citep{belkhale2023data}. Such behavior can introduce substantial noise and turns into the recorded trajectories. Through training, the policy learns these features and reproduces them in its predicted action chunks. At inference, these features provide an indication of whether the robot is at a decisive moment, how much error that phase can tolerate, and how far the execution rate can be adjusted.

Figure~\ref{fig:demo_phase}b visualizes Equation (5), showing how the rescaling factor varies with $I_{\cos}$ under the $2\times$ and $0.5\times$ commands. Under acceleration, the factor is large when the chunk follows a straight trajectory and decreases toward $1.0$ as CI increases. Under deceleration, the factor is close to the commanded stride when CI is small and increases as CI grows. In both cases, increasing CI brings the executed step lengths closer to those originally predicted by the policy.

\subsection{Evaluation in Simulation}

\label{sec:sim_fixed_exec}

We compare the following methods. (1) \textbf{\method{}} applies Equation~\eqref{eq:factor_mapping}. (2) \textbf{Vanilla resampling} is a training-free baseline used in prior works \citep{speedup_patch,race,demospeedup,speedtuning,espada}. When the policy predicts absolute end-effector poses or joint positions, this baseline simply resamples the actions and sends them to the robot for execution. In our delta-action setting, the equivalent operation sums the skipped action increments into the retained action. (3) \textbf{w/o CR} combines resampling with naive rescaling by the commanded stride, formally $f_t=\rho_t$, and serves as an ablation of \method{} without CI-based rescaling. All methods use the same frozen checkpoints and task initial states, with the unchanged $1\times$ policy as the reference. Under deceleration, vanilla resampling and w/o CR are equivalent, so only w/o CR is reported.

We run 50 rollouts per simulated task and report success rate, mean completion steps over successful trials, and realized rate. The realized rate is the ratio of the reference's mean completion steps to those of the evaluated method, computed only over trials in which both succeed. Flow Matching is evaluated at $\{0.3,\,0.5,\,1,\,1.5,\,2\}\times$, and $\pi_{0.5}$ and Fast-WAM at $\{0.2,\,0.3,\,0.5,\,1,\,2,\,3,\,4\}\times$. Table~\ref{tab:sim_fixed} reports the Flow Matching results. Across the four non-reference strides, \method{} maintains or exceeds the $1\times$ reference's mean success rate of $89$\%, with realized rates ranging from $0.348\times$ to $1.431\times$ across individual tasks. In comparison, both vanilla resampling and w/o CR achieve lower success rates than the $1\times$ reference.

\begin{figure}[tb]
  \centering
  \includegraphics[width=0.97\textwidth]{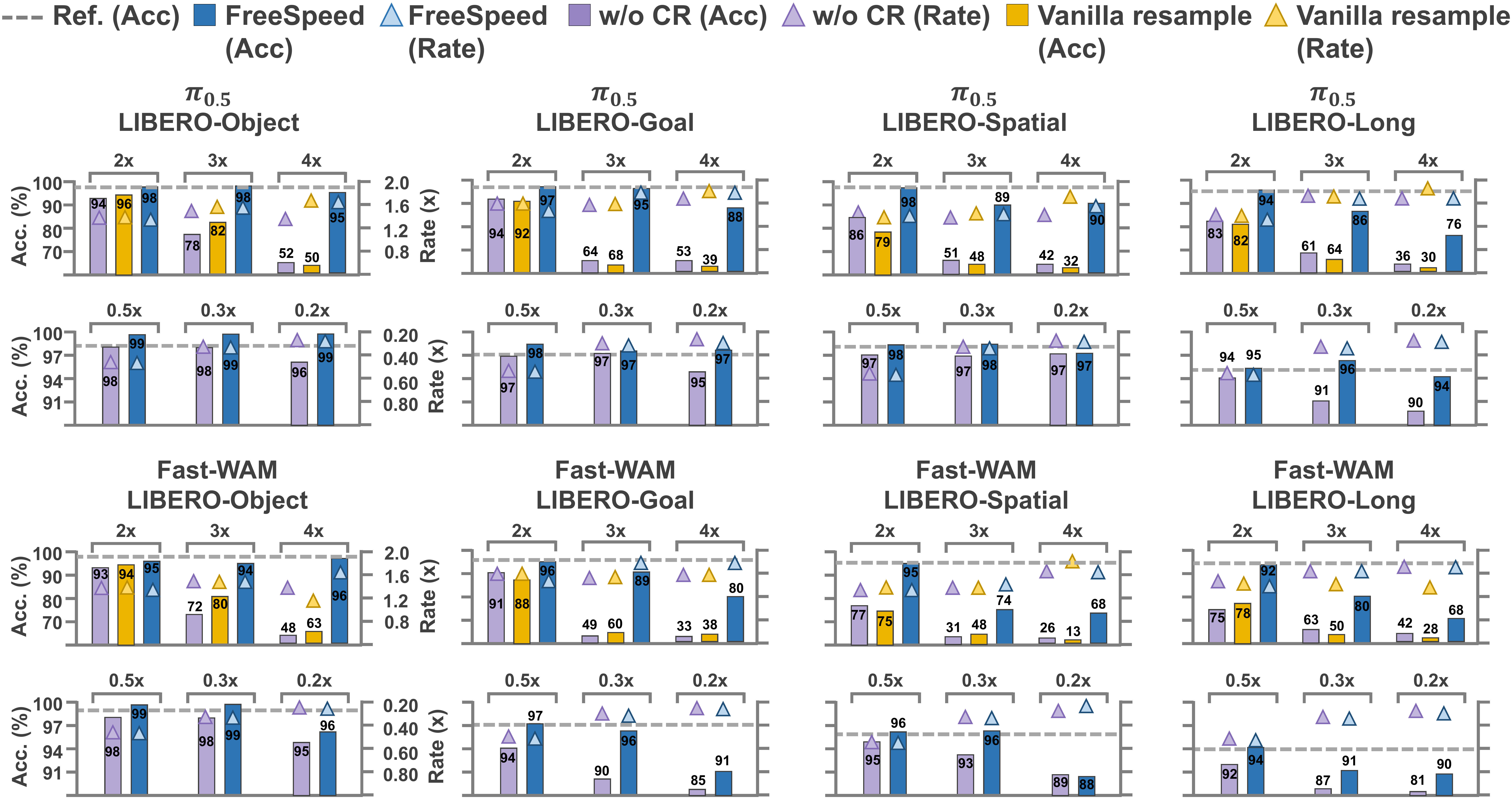}

  \caption{\textbf{Simulation comparison for $\pi_{0.5}$ and Fast-WAM.} Bars show success rates on the left axes, and triangular markers show realized execution rates on the right axes. The rate axes are inverted in the slowdown panels.}
  \label{fig:sim_comparison}
\end{figure}

Figure~\ref{fig:sim_comparison} extends the comparison to $\pi_{0.5}$ and Fast-WAM across the four LIBERO suites, each containing 10 tasks. At a $2\times$ command, \method{} maintains the reference success rates for both policies, whereas vanilla resampling reduces success by $9.2$ and $12.7$ percentage points and w/o CR by $8.0$ and $12.3$ percentage points, respectively. Under deceleration, \method{} outperforms w/o CR at every slowdown stride on both policies. Overall, across slowdown and speedup commands ranging from $0.2\times$ to $4\times$, \method{} achieves higher success rates than both baselines at every evaluated stride. Across individual tasks from both policies, the realized rates that preserve each task's own $1\times$ success rate range from $0.22\times$ to $2.53\times$.

\long\def\detailedfmsimtable{%
\begin{table}[!t]
\centering
\setlength{\belowcaptionskip}{4pt}
\caption{\textbf{Fixed-stride execution on 10 LIBERO-90 tasks with Flow Matching.} Success rates are reported in \%, with task indices defined in Table~\ref{tab:sim_task_list}. `Rate' is relative to the $1\times$ reference and averages per-task realized rates, each computed over trials in which both the evaluated method and the reference succeed. Bold indicates the highest success rate under each command. Completion steps are reported in Table~\ref{tab:sim_fixed_steps}.}
\label{tab:sim_fixed}
\scriptsize
\renewcommand{\arraystretch}{0.9}
\setlength{\tabcolsep}{2pt}
\begin{tabularx}{\textwidth}{@{}>{\raggedright\arraybackslash}p{40pt} >{\raggedright\arraybackslash}p{22pt} !{\vrule width 0.45pt} >{\columncolor{SimLavender}}c !{\vrule width 0.45pt} Y Y >{\columncolor{RRCBlue}}Y !{\vrule width 0.35pt} Y Y >{\columncolor{RRCBlue}}Y !{\vrule width 0.35pt} Y >{\columncolor{RRCBlue}}Y !{\vrule width 0.35pt} Y >{\columncolor{RRCBlue}}Y@{}}
\toprule
\multirow{2}{*}{Type} & \multirow{2}{*}{Task} & \multicolumn{1}{!{\vrule width 0.45pt}c!{\vrule width 0.45pt}}{$1\times$} & \multicolumn{3}{c!{\vrule width 0.35pt}}{$1.5\times$} & \multicolumn{3}{c!{\vrule width 0.35pt}}{$2\times$} & \multicolumn{2}{c!{\vrule width 0.35pt}}{$0.5\times$} & \multicolumn{2}{c}{$0.3\times$} \\
\cmidrule(lr){4-6}\cmidrule(lr){7-9}\cmidrule(lr){10-11}\cmidrule(lr){12-13}
& & Ref. & \wocrstack & \vanillastack & \methodstack & \wocrstack & \vanillastack & \methodstack & \wocrstack & \methodstack & \wocrstack & \methodstack \\
\midrule
\multirow{4}{40pt}{\raggedright Structure\\constrained} & 2 & $\mathbf{96}$ & $72$ & $82$ & $\mathbf{94}$ & $28$ & $36$ & $\mathbf{88}$ & $74$ & $\mathbf{92}$ & $46$ & $\mathbf{90}$ \\
& 32 & $\mathbf{78}$ & $\mathbf{76}$ & $66$ & $\mathbf{76}$ & $52$ & $50$ & $\mathbf{76}$ & $58$ & $\mathbf{70}$ & $66$ & $\mathbf{76}$ \\
& 40 & $\mathbf{90}$ & $70$ & $\mathbf{90}$ & $88$ & $78$ & $78$ & $\mathbf{88}$ & $\mathbf{100}$ & $94$ & $90$ & $\mathbf{96}$ \\
& 86 & $\mathbf{92}$ & $70$ & $84$ & $\mathbf{92}$ & $40$ & $64$ & $\mathbf{88}$ & $82$ & $\mathbf{90}$ & $\mathbf{86}$ & $\mathbf{86}$ \\
\midrule
\multirow{3}{40pt}{\raggedright Bounded\\receptacle} & 18 & $\mathbf{98}$ & $92$ & $96$ & $\mathbf{98}$ & $82$ & $96$ & $\mathbf{98}$ & $92$ & $\mathbf{98}$ & $\mathbf{96}$ & $\mathbf{96}$ \\
& 48 & $\mathbf{92}$ & $82$ & $82$ & $\mathbf{94}$ & $46$ & $90$ & $\mathbf{98}$ & $92$ & $\mathbf{98}$ & $96$ & $\mathbf{100}$ \\
& 60 & $\mathbf{100}$ & $86$ & $90$ & $\mathbf{100}$ & $60$ & $42$ & $\mathbf{100}$ & $86$ & $\mathbf{94}$ & $60$ & $\mathbf{94}$ \\
\midrule
\multirow{3}{40pt}{\raggedright Open\\relational} & 34 & $\mathbf{98}$ & $80$ & $82$ & $\mathbf{100}$ & $40$ & $76$ & $\mathbf{100}$ & $\mathbf{96}$ & $\mathbf{96}$ & $90$ & $\mathbf{96}$ \\
& 65 & $\mathbf{64}$ & $34$ & $42$ & $\mathbf{64}$ & $4$ & $16$ & $\mathbf{56}$ & $\mathbf{82}$ & $78$ & $74$ & $\mathbf{84}$ \\
& 69 & $\mathbf{80}$ & $68$ & $82$ & $\mathbf{92}$ & $28$ & $64$ & $\mathbf{94}$ & $74$ & $\mathbf{88}$ & $38$ & $\mathbf{80}$ \\
\midrule
\multicolumn{2}{@{}l}{Mean} & $\mathbf{89}$ & $73$ & $80$ & $\mathbf{90}$ & $46$ & $61$ & $\mathbf{89}$ & $84$ & $\mathbf{90}$ & $74$ & $\mathbf{90}$ \\
\multicolumn{2}{@{}l}{Rate ($\times$)} & $1.000$ & $1.242$ & $1.280$ & $1.221$ & $1.435$ & $1.483$ & $1.353$ & $0.570$ & $0.626$ & $0.328$ & $0.402$ \\
\bottomrule
\end{tabularx}

\end{table}

}

\detailedfmsimtable

\begin{table}[!t]

\centering
\setlength{\belowcaptionskip}{8pt}
\caption{\textbf{Fixed-stride execution on four real-robot tasks.} Per-task success rates are reported in \%. Bold indicates the highest success rate under each command. Per-task completion steps and realized rates are reported in Table~\ref{tab:real_fixed_steps}.}
\label{tab:real_fixed}
\scriptsize
\renewcommand{\arraystretch}{0.9}
\setlength{\tabcolsep}{0.9pt}
\begin{tabularx}{\textwidth}{@{}>{\raggedright\arraybackslash}p{46pt} !{\vrule width 0.45pt} >{\columncolor{SimLavender}}Y !{\vrule width 0.45pt} Y Y >{\columncolor{RRCBlue}}Y !{\vrule width 0.35pt} Y Y >{\columncolor{RRCBlue}}Y !{\vrule width 0.35pt} Y Y >{\columncolor{RRCBlue}}Y !{\vrule width 0.35pt} Y >{\columncolor{RRCBlue}}Y !{\vrule width 0.35pt} Y >{\columncolor{RRCBlue}}Y !{\vrule width 0.35pt} Y >{\columncolor{RRCBlue}}Y@{}}
\toprule
 & \multicolumn{1}{c!{\vrule width 0.45pt}}{$1\times$} & \multicolumn{3}{c!{\vrule width 0.35pt}}{$2\times$} & \multicolumn{3}{c!{\vrule width 0.35pt}}{$3\times$} & \multicolumn{3}{c!{\vrule width 0.35pt}}{$4\times$} & \multicolumn{2}{c!{\vrule width 0.35pt}}{$0.5\times$} & \multicolumn{2}{c!{\vrule width 0.35pt}}{$0.3\times$} & \multicolumn{2}{c}{$0.2\times$} \\
\cmidrule(lr){2-2}\cmidrule(lr){3-5}\cmidrule(lr){6-8}\cmidrule(lr){9-11}\cmidrule(lr){12-13}\cmidrule(lr){14-15}\cmidrule(lr){16-17}
 & Ref. & \wocrstack & \vanillastack & \methodstack & \wocrstack & \vanillastack & \methodstack & \wocrstack & \vanillastack & \methodstack & \wocrstack & \methodstack & \wocrstack & \methodstack & \wocrstack & \methodstack \\
\midrule
Peach on Plate & $100$ & $70$ & $90$ & $\mathbf{100}$ & $50$ & $55$ & $\mathbf{100}$ & $45$ & $0$ & $\mathbf{95}$ & $70$ & $\mathbf{100}$ & $65$ & $\mathbf{100}$ & $20$ & $\mathbf{95}$ \\
Tennis in Can & $90$ & $90$ & $60$ & $\mathbf{95}$ & $45$ & $30$ & $\mathbf{90}$ & $0$ & $0$ & $\mathbf{85}$ & $80$ & $\mathbf{95}$ & $70$ & $\mathbf{90}$ & $65$ & $\mathbf{95}$ \\
Stack cube & $90$ & $60$ & $75$ & $\mathbf{95}$ & $0$ & $20$ & $\mathbf{90}$ & $5$ & $0$ & $\mathbf{90}$ & $\mathbf{95}$ & $90$ & $65$ & $\mathbf{95}$ & $\mathbf{75}$ & $\mathbf{75}$ \\
Pour almond & $95$ & $70$ & $85$ & $\mathbf{100}$ & $55$ & $50$ & $\mathbf{100}$ & $50$ & $45$ & $\mathbf{95}$ & $85$ & $\mathbf{100}$ & $65$ & $\mathbf{90}$ & $60$ & $\mathbf{95}$ \\
\midrule
Mean & $93.8$ & $72.5$ & $77.5$ & $\mathbf{97.5}$ & $37.5$ & $38.8$ & $\mathbf{95.0}$ & $25.0$ & $11.2$ & $\mathbf{91.2}$ & $82.5$ & $\mathbf{96.2}$ & $66.2$ & $\mathbf{93.8}$ & $55.0$ & $\mathbf{90.0}$ \\
Rate ($\times$) & $1.00$ & $1.66$ & $1.69$ & $1.44$ & $1.86$ & $1.93$ & $1.62$ & $1.93$ & $2.18$ & $1.75$ & $0.57$ & $0.68$ & $0.37$ & $0.47$ & $0.25$ & $0.40$ \\
\bottomrule
\end{tabularx}

\end{table}

\subsection{Evaluation on the Real Robot}

\label{sec:real_robot}

We apply \method{} in the same plug-and-play manner to pretrained flow-matching policies for four real-robot tasks, with each policy trained on $50$ demonstrations. A Rokae Helios series humanoid robot executes the action at $30$ Hz. The commands are $\{0.2,\,0.3,\,0.5,\,1,\,2,\,3,\,4\}\times$, and Table~\ref{tab:real_fixed} reports the results for all four tasks. Further details are provided in Appendix~\ref{app:dataset_real}.

Across the four tasks, the $1\times$ reference achieves a mean success rate of $93.8$\%. Under \method{} commands of $0.2\times$, $0.3\times$, $0.5\times$, $2\times$, $3\times$, and $4\times$, the mean success rates are $90.0$\%, $93.8$\%, $96.2$\%, $97.5$\%, $95.0$\%, and $91.2$\%, respectively. In contrast, at acceleration commands of $2\times$, $3\times$, and $4\times$, vanilla resampling achieves $77.5$\%, $38.8$\%, and $11.2$\% mean success, while w/o CR achieves $72.5$\%, $37.5$\%, and $25.0$\%, respectively. Averaged over the six non-reference strides, \method{} achieves $94.0$\% mean success, compared with the reference's $93.8$\%, with realized execution rates ranging from $0.38\times$ to $1.97\times$ across individual tasks.

\subsection{\method{} Keeps the Policy In Distribution}

\label{sec:in_distribution}

\begin{figure}[!t]
  \centering
  \includegraphics[width=0.97\textwidth]{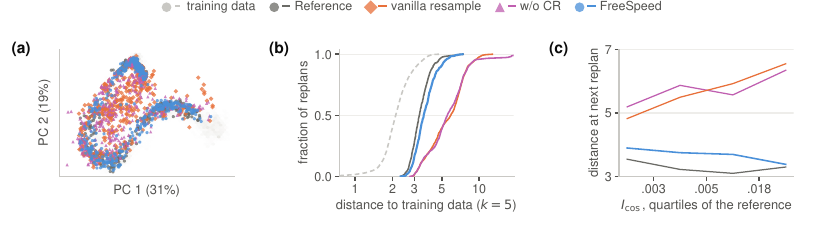}

  \caption{\textbf{Observation distribution shift under the $2\times$ command.} \method{} remains close to the $1\times$ reference in (a) and (b), while the baselines show larger deviations from the training data, particularly after executing chunks with high cosine inconsistency in (c).}

  \label{fig:ood_grid}
\end{figure}

To examine how rate modification affects the policy's input distribution, we analyze Flow Matching on LIBERO-90 Task 60. Figure~\ref{fig:ood_grid}a projects observation embeddings from training demonstrations and execution rollouts onto two principal axes fitted to the training data. Figure~\ref{fig:ood_grid}b plots empirical cumulative distributions of each execution embedding's mean Euclidean distance to its $k=5$ nearest training embeddings in the full observation conditioning space \citep{knn_ood,knn_ood_highdim}. Appendix~\ref{app:ood_full} provides the sampling and measurement details of the above plots. At $2\times$, \method{} remains close to the $1\times$ reference and training data, with a median distance of $1.09$ times the reference median, compared with $1.79$ for vanilla resampling and $1.68$ for w/o CR. The slowdown analysis at $0.3\times$ shows the same trend (Figure~\ref{fig:ood_slow}), indicating that \method{} limits observation shifts under both acceleration and deceleration.

Figure~\ref{fig:ood_grid}c relates each chunk's $I_{\cos}$ to the distance from the next replan's observation embedding to the training data, using the same $k=5$ nearest-neighbor measure. Under the two baselines, higher inconsistency is associated with larger subsequent distances, whereas \method{} stays close to the reference across the inconsistency range. This supports the intuition that changing step lengths is more disruptive around turns and noisy motion than along straight segments, motivating less aggressive rescaling when predicted directions become inconsistent.

\subsection{Rate That Varies During Execution}

The preceding evaluations use fixed-stride FreeSpeed: the resampling stride remains fixed throughout each task rollout, and each chunk uses the rescaling bound determined by Equation~\ref{eq:factor_mapping}. FreeSpeed also supports more flexible execution settings. We now demonstrate that it maintains task success under two conditions: stride randomization across policy inferences and factor randomization within FreeSpeed's bounds. Table~\ref{tab:sim_dynamic} reports both cases. For stride randomization, we sample a new stride from the full command set of $\{0.3,\,0.5,\,1,\,1.5,\,2\}\times$ for flow matching and $\{0.2,\,0.3,\,0.5,\,1,\,2,\,3,\,4\}\times$ for $\pi_{0.5}$ at each replan. Success rates reach $90$\% for Flow Matching and $97$\% for $\pi_{0.5}$, comparable to their respective references. For factor randomization, we keep the stride fixed and compute the rescaling bound $f_t$ following Equation~\ref{eq:factor_mapping}. Instead of using $f_t$ directly, as in the preceding evaluations, we sample a factor between $f_{\mathrm{c}}$ and $f_t$ at each replan. Across the four evaluated strides on $\pi_{0.5}$ and Flow Matching, success rates remain within 2 percentage points of those obtained with the fixed-stride rule.

\begin{table}[!t]
\centering
\setlength{\belowcaptionskip}{4pt}
\caption{\textbf{Within-episode stride and factor randomization.} FM is evaluated on the $10$ tasks in Table~\ref{tab:sim_fixed}, and $\pi_{0.5}$ on all $40$ tasks in Figure~\ref{fig:sim_comparison}. (A) A new stride is sampled at each replan. (B) The stride remains fixed, while the rescaling factor is sampled between $f_{\mathrm{c}}$ and $f_t$.}
\label{tab:sim_dynamic}
\scriptsize
\renewcommand{\arraystretch}{0.9}
\setlength{\tabcolsep}{3pt}
\begin{tabularx}{\textwidth}{@{}l l !{\vrule width 0.45pt} Y !{\vrule width 0.3pt} Y !{\vrule width 0.45pt} YYYY !{\vrule width 0.3pt} YYYY@{}}
\toprule
& & \multicolumn{2}{c!{\vrule width 0.45pt}}{\bfseries A: the stride varies}
& \multicolumn{8}{c}{\bfseries B: the factor varies} \\
\cmidrule(lr){3-4}\cmidrule(lr){5-12}
& & FM & $\pi_{0.5}$
& \multicolumn{4}{c!{\vrule width 0.3pt}}{FM}
& \multicolumn{4}{c}{$\pi_{0.5}$} \\
\cmidrule(lr){3-3}\cmidrule(lr){4-4}\cmidrule(lr){5-8}\cmidrule(lr){9-12}
& & dynamic & dynamic
& $0.3\times$ & $0.5\times$ & $1.5\times$ & $2\times$
& $0.3\times$ & $0.5\times$ & $1.5\times$ & $2\times$ \\
\midrule
\method{} & Succ. & -- & -- & $90$ & $90$ & $90$ & $89$ & $97$ & $98$ & $97$ & $97$ \\
& Rate & -- & -- & $0.40$ & $0.63$ & $1.22$ & $1.35$ & $0.39$ & $0.57$ & $1.21$ & $1.34$ \\
\rowcolor{RRCBlue}\cellcolor{white}varying & \cellcolor{white}Succ.
& $90$ & $97$ & $89$ & $91$ & $92$ & $91$ & $97$ & $98$ & $96$ & $97$ \\
\rowcolor{RRCBlue}\cellcolor{white} & \cellcolor{white}Rate
& $0.63$ & $0.54$ & $0.46$ & $0.68$ & $1.20$ & $1.28$ & $0.42$ & $0.61$ & $1.13$ & $1.21$ \\
\midrule
Reference & \multicolumn{11}{c@{}}{\cellcolor{SimLavender}
FM: $88$\% success at $1.00\times$. $\pi_{0.5}$: $97$\% success at $1.00\times$.} \\
\bottomrule
\end{tabularx}
\end{table}

\begin{figure}[!t]

  \centering
  \includegraphics[width=0.98\textwidth]{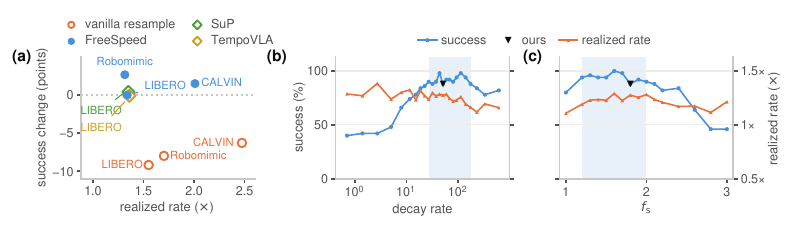}

  \caption{\textbf{Comparison with published methods and sensitivity to the two constants.} \textbf{(a)} Success change against realized rate. \textbf{(b, c)} The decay rate $\lambda$ and the factor $f_{\mathrm{s}}$ swept in turn.}

  \label{fig:fixed_rate_curves}\label{fig:sensitivity}\label{fig:benchmark_panels}
\end{figure}

\subsection{Comparison with Prior Methods and Ablation}

\label{sec:comparison}

We compare \method{} against SuP \citep{speedup_patch}, TempoVLA \citep{tempovla}, and vanilla resampling on the LIBERO benchmark, using the success rates reported in the original SuP and TempoVLA papers. We also evaluate \method{} on other benchmarks, including CALVIN and Robomimic \citep{calvin, robomimic} (Appendix~\ref{app:CALVIN_details}). In Figure~\ref{fig:benchmark_panels}a, the horizontal axis shows the realized rate, and the vertical axis shows the change in success rate relative to the reference. We report this change rather than the absolute success rate because differences in random seeds and other evaluation settings make direct alignment difficult, making relative changes more informative. On LIBERO, \method{} matches the compared methods in success-rate change and realized rate while providing a training-free, plug-and-play solution. Vanilla resampling can achieve higher realized rates, but at the cost of a substantial reduction in success rate.

The w/o CR method serves as an ablation, with simulation and real-robot results already reported in Sections~\ref{sec:sim_fixed_exec} and~\ref{sec:real_robot}. Figure~\ref{fig:real_curves} in the appendix plots these results, making it easier to identify the commanded rate at which each rule no longer preserves the reference success rate. 

We further assess whether the adaptive design of \method{} is necessary by applying a single rescaling factor throughout execution. This ablation is similar to w/o CR but uses a deliberately smaller factor instead of setting it equal to the commanded stride. It tests whether simply trading speed for success can recover the w/o CR baseline's performance. Appendix~\ref{app:constant_factor} sweeps the constant factor from $1.0$ to $2.0$ at the $2\times$ command and reports the result in full.

We also investigate parameter sensitivity. We assess sensitivity to the decay rate $\lambda$ and the factor $f_{\mathrm{s}}$. Figure~\ref{fig:sensitivity}b and c vary each constant in turn at the $2\times$ command, with $50$ rollouts per setting. The left and right axes show success rate and realized rate, respectively. Shading indicates the range of settings whose success rates remain within ten percentage points of the best observed result.

\section{Conclusion}

\label{sec:conclusion}

\method{} enables test-time execution-speed control for frozen action-chunking policies. It computes cosine inconsistency between adjacent actions within each predicted chunk and uses this signal to adapt rescaling under fixed or time-varying commands. Evaluations with Flow Matching, $\pi_{0.5}$, and Fast-WAM on 10 LIBERO-90 tasks and the four LIBERO suites demonstrate speed adjustment over a range of realized rates while maintaining success rates comparable to the frozen-policy references. Across four real-robot tasks, \method{} likewise matches the frozen policy's mean success rate when averaged over six commands from $0.2\times$ to $4\times$. The main limitation of \method{} is that its acceleration ceiling is set by how much free motion a task contains, as detailed in Appendix~\ref{app:limitations}.

\section*{AI Use Statement}
Large Language Models (LLMs) were used to assist with the writing and polishing of the manuscript, including sentence rephrasing, grammar checking, and improving clarity and readability. LLMs were not involved in research ideation, methodology, experimental design, or data analysis. All AI-assisted text was reviewed and revised by the authors, who take full responsibility for the final content of the manuscript.

\section*{Ethics Statement}
This work adheres to the ICLR Code of Ethics. Our experiments do not involve human subjects or sensitive personal data.

\begin{minipage}{\linewidth}
\section*{Reproducibility Statement}
The complete algorithm and implementation details of FreeSpeed are described in the main text. All implementation parameters, experimental settings, and robot hardware specifications are provided in the Appendix to support the replicability of our experiments.
\end{minipage}

\bibliographystyle{assets/plainnat}
\bibliography{refs}

\clearpage
\beginappendix
\setcounter{table}{0}\renewcommand{\thetable}{A\arabic{table}}
\setcounter{figure}{0}\renewcommand{\thefigure}{A\arabic{figure}}
\setcounter{equation}{0}\renewcommand{\theequation}{A\arabic{equation}}
\renewcommand{\theHtable}{A\arabic{table}}
\renewcommand{\theHfigure}{A\arabic{figure}}
\renewcommand{\theHequation}{A\arabic{equation}}

\section{Hardware Setup}
\label{app:hardware}

The real-robot experiments are conducted on a Rokae Helios series humanoid robot, using one of its two seven-degree-of-freedom arms equipped with a Daimon gripper. The arm operates in Cartesian impedance mode and executes policy-predicted end-effector increments at $30$ Hz. We collect $50$ demonstrations per task through teleoperation with a PICO controller and perform $20$ rollouts for each evaluated condition. Table~\ref{tab:hardware} lists the hardware and data collection settings.

\begin{figure}[!htbp]
  \centering
  \includegraphics[width=0.5\textwidth]{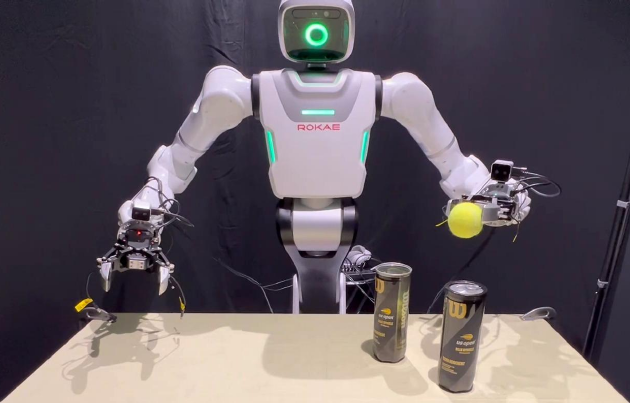}
  \caption{\textbf{The real-robot workstation.} The experiments use the robot's left arm, shown on the right in the image. This frame is taken from a $1\times$ rollout of the tennis-ball task, with the ball in the gripper and two cans on the table.}
  \label{fig:hardware_setup}
\end{figure}

\begin{table}[!htbp]
\centering
\setlength{\belowcaptionskip}{4pt}
\caption{\textbf{Real-robot hardware and demonstration collection.}}
\label{tab:hardware}
\small
\renewcommand{\arraystretch}{1.45}
\begin{tabular}{@{}p{0.30\textwidth}@{\hspace{1.5em}}p{0.42\textwidth}@{}}
\toprule
Item & Value \\
\midrule
Robot & Rokae Helios series humanoid \\
 Arm & 7 DoF \\
End effector & Daimon gripper \\
robot command rate & $30$ Hz \\
Action interface & Cartesian end-effector increments \\
Head camera & Orbbec, $640\times480$, $30$ fps \\
Wrist camera & Intel RealSense, $640\times480$, $60$ fps \\
Teleoperation & PICO controller \\
Demonstration rate & $30$ fps \\
Demonstrations per task & $50$ \\
Rollouts per condition & $20$ \\
\bottomrule
\end{tabular}
\end{table}

Depth images and fingertip tactile readings are recorded during data collection, but the trained policy uses only the two RGB streams from the wrist and head cameras and the proprioceptive state.

\section{Dataset Details}
\label{app:dataset}

\subsection{Simulation Tasks}
\label{app:dataset_sim}

The flow-matching evaluation uses 10 tasks drawn from LIBERO-90, listed in Table~\ref{tab:sim_task_list}. Each task provides 50 human demonstrations, and we train one policy per task. All 10 are pick-and-place tasks, sharing two critical phases: grasping the object and releasing it at the target. These shared phases provide a common basis for examining the effects of speed modification. The tasks differ in their placement constraints, particularly the allowable variation in release position, and we group them into three types accordingly.

\paragraph{Structure-constrained placement.} The target is a fixed part of the scene, such as an open drawer or a cabinet shelf. The release position allows only a few centimetres of deviation, and the approach must clear the surrounding geometry. Overshooting during accelerated execution can cause the object or gripper to collide with nearby structures. We therefore expect these tasks to be particularly sensitive to speed increases during the approach.

\paragraph{Bounded receptacle placement.} The target is a container or a flat region with an edge, such as a basket, a tray or a stovetop, that accepts the object within a limited area. The release position allows several centimetres of variation in the horizontal plane, but the object must still clear the rim or that edge before release. These tasks permit greater variation in placement than the first group but remain sensitive to motion near the boundary and during release.

\paragraph{Open relational placement.} The target is specified relative to another object on an open surface, such as on a plate or to its left. Without enclosing geometry, these tasks allow the greatest placement tolerance among the three groups. We therefore expect placement to be less sensitive to speed modification, while grasping remains a critical phase.

This grouping is specific to our analysis and reflects how placement constraints may affect tolerance to speed modification. Table~\ref{tab:sim_fixed} groups the tasks accordingly, allowing the effects of fixed-stride execution to be compared across different levels of placement tolerance.

\begin{table}[!htbp]
\centering
\setlength{\belowcaptionskip}{8pt}
\caption{\textbf{The 10 LIBERO-90 tasks used for the flow-matching evaluation.} Each task is listed with its LIBERO-90 index and corresponding language instruction.}
\label{tab:sim_task_list}
\small
\renewcommand{\arraystretch}{1.05}
\begin{tabularx}{\textwidth}{@{}p{66pt} p{24pt} X@{}}
\toprule
Type & Task & Language instruction \\
\midrule
\multirow{4}{66pt}{\raggedright Structure constrained} & 2 & put the black bowl in the top drawer of the cabinet \\
 & 32 & put the ketchup in the top drawer of the cabinet \\
 & 40 & put the frying pan on the cabinet shelf \\
 & 86 & pick up the book in the middle and place it on the cabinet shelf \\
\midrule
\multirow{3}{66pt}{\raggedright Bounded receptacle} & 18 & put the frying pan on the stove \\
 & 48 & pick up the ketchup and put it in the basket \\
 & 60 & pick up the black bowl on the left and put it in the tray \\
\midrule
\multirow{3}{66pt}{\raggedright Open relational} & 34 & put the yellow and white mug to the front of the white mug \\
 & 65 & put the red mug on the left plate \\
 & 69 & put the chocolate pudding to the left of the plate \\
\bottomrule
\end{tabularx}
\end{table}

\subsection{Real-World Tasks}
\label{app:dataset_real}

The four real-robot tasks are: (1) placing a peach on a plate, (2) dropping a tennis ball into a bin, (3) stacking one cube on another, and (4) pouring almonds from one cup into another. The policy for each task is trained on $50$ teleoperated demonstrations.

\section{Implementation Details}
\label{app:implementation}

\subsection{Frozen Policies}
\label{app:impl_policies}

We evaluate \method{} on three frozen action-chunking policies. We train one Flow Matching model \citep{pmlr-v270-chisari25a,zhang2025flowpolicy,bai2026flash} per task using the task's original demonstrations, without speed-related processing. For $\pi_{0.5}$ and Fast-WAM, we use the released checkpoints without additional training or fine-tuning \citep{pi05,fastwam}. All policy weights remain frozen throughout evaluation, and \method{} operates only on their predicted action chunks.

The Flow Matching policy is evaluated on 10 LIBERO-90 tasks and four real-robot tasks, with $50$ training demonstrations per task. $\pi_{0.5}$ and Fast-WAM are evaluated on four LIBERO suites: LIBERO-Spatial, LIBERO-Object, LIBERO-Goal, and LIBERO-Long, each containing 10 tasks. Table~\ref{tab:policies} lists the settings for all three policies.

\begin{table}[!htbp]
\centering
\setlength{\belowcaptionskip}{4pt}
\caption{\textbf{Configurations of the three frozen policies.} Execution horizons are reported for the original $1\times$ setting. Demonstration counts are provided only for the Flow Matching policies trained in this work; dashes indicate counts not listed for the released checkpoints.}
\label{tab:policies}
\small
\renewcommand{\arraystretch}{1.05}
\begin{tabularx}{\textwidth}{@{}l X X X@{}}
\toprule
 & Flow Matching & $\pi_{0.5}$ & Fast-WAM \\
\midrule
Source & trained in this work & released checkpoint & released checkpoint \\
Observation history & $8$ frames & $1$ frame & $1$ frame \\
Steps executed per replan & $12$ & $8$ & $10$ \\
Evaluation tasks & 10 LIBERO-90 and 4 real & 4 LIBERO suites, 10 tasks each & 4 LIBERO suites, 10 tasks each \\
Demonstrations per task & $50$ & -- & -- \\
\bottomrule
\end{tabularx}
\end{table}

Since we train the base flow-matching models from scratch, we provide their architecture and training details below. The policy takes eight consecutive observations and predicts twelve seven-dimensional actions. Each observation includes two $128\times128$ RGB images, joint positions, end-effector state, gripper state, and a language instruction. Frozen BERT-base-cased features \citep{devlin-etal-2019-bert,liu2025rdt} are attention-pooled into a 64-dimensional language vector. Each camera uses a separate truncated ResNet-18 encoder \citep{He_2016_CVPR} with language-conditioned FiLM \citep{perez2018film,brohan2022rt,pmlr-v164-jang22a,haldar2024baku,kim2025fine} and spatial-softmax pooling \citep{JMLR:v17:15-522}. Each of the six modalities is encoded into 64 dimensions, and the features are concatenated across the observation history to form a $3072$-dimensional conditioning vector. A conditional one-dimensional U-Net \citep{ronneberger2015u} with channel widths $(256,512,1024)$ generates actions. The action-generation branch contains approximately $111.1$ million trainable parameters.

Training uses linear Gaussian-to-action paths and a mean-squared-error loss on the predicted vector field. The documented configuration uses batch size $128$ and AdamW \citep{loshchilov2018decoupled} with learning rate $10^{-4}$, weight decay $10^{-4}$, and cosine learning-rate decay \citep{loshchilov2017sgdr} to $10^{-5}$. Inference uses ten explicit Euler steps. For the real-robot checkpoint, generating one action chunk takes $22.9\pm0.45$ ms on an NVIDIA RTX 5090 at batch size one using FP32. This timing includes observation encoding but excludes preprocessing, communication, and BERT computation, as BERT features are precomputed and cached. Deployment logs separately report an average server inference time of $33.2$ ms.

\subsection{\method{}}
\label{app:impl_method}

For each commanded rate $\rho_t$, we fix two scaling endpoints, $f_{\mathrm{s}}(\rho_t)$ and $f_{\mathrm{c}}(\rho_t)$. The factor equals $f_{\mathrm{s}}$ when cosine inconsistency is zero and approaches $f_{\mathrm{c}}$ as inconsistency increases. Under acceleration, $f_{\mathrm{c}}=1$, so increasing inconsistency brings the executed step lengths closer to the original predictions while retaining temporal resampling. Under deceleration, $f_{\mathrm{s}}=\rho_t$, giving the full slowdown scaling for consistent chunks, whereas increasing inconsistency moves the factor toward $f_{\mathrm{c}}$, which is closer to $1$. Table~\ref{tab:factor_endpoints} lists all endpoint values used in this work. These values depend only on the commanded rate and are shared across tasks, policies, and rollouts, without task-specific tuning. For all acceleration commands, we set $f_{\mathrm{s}}(\rho_t)=0.9\rho_t$.

We use a single decay parameter, $\lambda=50$, across all commands and policies.

\begin{table}[h]
\centering
\setlength{\belowcaptionskip}{4pt}
\caption{\textbf{Scaling endpoints.} One pair per commanded rate, shared by
every policy and both simulation and the real robot.}
\label{tab:factor_endpoints}
\small
\renewcommand{\arraystretch}{1.0}
\begin{tabular}{@{}lcccccccc@{}}
\toprule
$\rho$ & $0.2$ & $0.3$ & $0.5$ & $1$ & $1.5$ & $2$ & $3$ & $4$ \\
\midrule
$f_{\mathrm{s}}$ & $0.20$ & $0.30$ & $0.50$ & $1.00$ & $1.35$ & $1.80$ & $2.70$ & $3.60$ \\
$f_{\mathrm{c}}$ & $0.35$ & $0.50$ & $0.80$ & $1.00$ & $1.00$ & $1.00$ & $1.00$ & $1.00$ \\
\bottomrule
\end{tabular}
\end{table}

\subsection{Controller Action Limits in Simulation}
\label{app:impl_scale_limit}

LIBERO controls the arm through robosuite's operational space controller \citep{1087068,robosuite,mujoco} using end-effector pose increments \citep{10522877}. The controller clips each action dimension independently to $[-1,1]$ and linearly maps it to translation increments of up to $0.05$ m or rotation increments of up to $0.5$ rad per component. Such element-wise clipping can alter both the magnitude and direction of an action, distorting scaled actions that exceed the bounds.

To avoid this distortion, we expand the input range to $[-L,L]$ and proportionally expand the output range. This preserves the end-effector increment per unit action while relaxing the clipping bound. All other controller settings remain at their LIBERO defaults. Since the largest $f_{\mathrm{s}}$ used in this work is $3.6$, we set $L=4$ for every rate-modified condition and retain the default $L=1$ for the $1\times$ reference. All methods use the same $L$ for a given commanded rate.

Under the fixed $2\times$ stride, $59.7$\% of actions exceed the original input bound ($\max_j|a_j|>1$). With $L=4$, no actions exceed the expanded bound, with the largest observed absolute action component being $1.52$. In contrast, the $1\times$ and $0.5\times$ conditions are nearly unaffected by the original bound. On the real robot, translation and rotation increments are limited to $35$ mm and $60$ mrad per step, respectively. No commands exceed these limits at the execution rates evaluated in this work.

\section{Comparison Criteria and Data Sources}
\label{app:published_numbers}

This section defines the criteria used in Table~\ref{tab:capabilities} and identifies the sources of the results plotted in Figure~\ref{fig:benchmark_panels}a.

\subsection{Criteria in the Capability Table}

Each column of Table~\ref{tab:capabilities} represents a binary criterion. A filled dot indicates that the capability is demonstrated in the paper; an open dot indicates that it is not demonstrated.

\paragraph{No rate training.} Whether the rate-control mechanism requires no training or fine-tuning. This criterion concerns the rate-control mechanism itself, rather than the entire policy pipeline. RACE satisfies it because its retiming uses classical time-optimal path parameterization \citep{toppra}, which requires no training. Although RACE retrains the policy to predict reached states rather than commanded states, this changes the policy's prediction target rather than introducing a learned rate-control mechanism.

\paragraph{Unmodified demos.} Whether the method uses the collected demonstrations without offline modification to support speed control, such as retiming, downsampling, segment annotation, or conversion to a different supervision target. Under this definition, AutoSpeed satisfies the criterion: it retimes the same future trajectory at several rates to generate candidate supervision targets, evaluates them using a composite cost, and trains toward the lowest-cost candidate. These operations occur within the training objective rather than through offline modification of the demonstration dataset.

\paragraph{Unchanged controller.} Whether the reported speed modification retains the low-level control law and controller gains used during demonstration collection. Controller gains and tracking accuracy affect how quickly the robot can follow motion commands. Replacing the original controller with a higher-gain controller can therefore make part of the reported speedup attributable to changes in low-level control.

\paragraph{User-set rate.} Whether the method accepts a user-specified target execution rate and adjusts execution accordingly. A rate inferred or optimized autonomously by the method does not satisfy this criterion.

\paragraph{Slow-motion.} Whether the method supports execution below the original $1\times$ rate. Methods restricted to skipping actions or compressing trajectories in time do not satisfy this criterion unless they also provide a mechanism for slowing execution.

\paragraph{In-episode rate.} Whether the user can update the target rate during an episode and have the new command take effect online within the method's supported range. This criterion concerns user control rather than internally varying execution rates. SuP's scheduler changes the downsampling rate between action chunks, but it does not accept a user-specified target rate and therefore does not satisfy this criterion.

\subsection{Method Selection in the Benchmark Comparison}

Figure~\ref{fig:benchmark_panels}a compares changes in success rate with realized execution rate, both relative to each method's own $1\times$ reference. Table~\ref{tab:capabilities} shows that existing methods differ from \method{} in their training requirements, demonstration processing, controller settings, support for user-specified rates, deceleration, and rate adjustment within an episode. Since no method matches the complete setting of \method{}, we select representative methods along two complementary dimensions: post-processing action chunks from a frozen policy and accepting user-specified execution rates.

We select SuP and TempoVLA to represent these two dimensions. SuP freezes the base policy and downsamples its predicted action chunks, making it the closest comparison in execution mechanism. It trains a scheduler through offline reinforcement learning using counterfactual trajectories generated by a learned world model. The scheduler selects rates autonomously, and SuP applies benchmark-specific gripper compensation through increased control gain or action magnitude. TempoVLA is the closest comparison in control interface: it accepts user-specified rates, supports acceleration and deceleration, and allows rate changes during execution. These capabilities, however, require a speed-conditioned policy trained on demonstrations retimed to multiple rates.

SAIL and RACE study execution beyond demonstration speed while addressing the associated tracking challenges. Their evaluations use high-gain controllers that differ from those used during demonstration collection, so their reported results reflect both rate modulation and changes in low-level control. The contribution of the controller is evident in SAIL's ablation: without high-gain control, the $-$HG configuration in Table J.6 of its appendix achieves an average realized rate of $0.94\times$ across five tasks, with success $19.2$ percentage points below the diffusion-policy reference. Neither method accepts a user-specified target rate. These differences in control conditions and rate specification make them less directly aligned with our comparison.

RACE similarly changes the low-level controller. It retrains the policy to predict reached states that a higher-gain controller tracks during execution, then applies TOPP-RA \citep{toppra} to retime the resulting state trajectory. TOPP-RA computes a time-optimal parameterization of a given geometric path under kinodynamic constraints while preserving its geometry. Applying TOPP-RA in our setting would require specifying how to construct the geometric path, define the kinodynamic constraints, and convert the retimed trajectory into control commands. Furthermore, when the original low-level controller does not tightly track commanded states, deviations between commanded and achieved motion can complicate the tracking of retimed end-effector or joint trajectories. Integrating TOPP-RA would therefore require additional optimization and tracking considerations, as addressed in RACE, and entail broader changes to the execution pipeline than FreeSpeed's direct action-chunk post-processing, which leaves the underlying control and policy inference unchanged.

The remaining methods primarily optimize execution efficiency through autonomous rate selection. DemoSpeedup processes demonstrations through entropy-guided downsampling, retrains the policy, and increases the gripper gain during evaluation. AutoSpeed constructs candidate targets at different rates and trains a Ratio Head to predict the rate. SpeedTuning freezes the base policy but trains an external speed policy through online reinforcement learning to interpolate its action chunks. SpeedAug constructs multi-rate targets through random downsampling and further optimizes the policy through online reinforcement learning. These methods infer or optimize the rate rather than accept it as a user command, so they do not directly support the same analysis of commanded rate, realized rate, and success rate as \method{}. We include SuP as a representative autonomous method because its post-processing of frozen-policy action chunks is most closely related to our mechanism, while TempoVLA provides the comparison in user control. Their differences in training, data processing, and controller settings remain relevant when interpreting the results.

\subsection{Metric Reporting}

We report changes in success rate relative to each paper's own $1\times$ reference, expressed in percentage points, to account for differences in performance across base models. The realized execution rate is either taken directly from the source or calculated from its reported mean episode lengths over successful rollouts, dividing the reference mean by the evaluated condition's mean.

\subsection{SuP}

We use the $\pi_{0.5}$ results in Table 2 of the original paper. Averaged over the four LIBERO suites, the base policy achieves $96.9$\% success at $1.00\times$, while SuP achieves $97.3$\% at $1.35\times$. These directly reported values correspond to a success-rate increase of $0.4$ percentage points.

\subsection{TempoVLA}

We use the text setting in Table 2 of the original paper. At the $1.0\times$ command, TempoVLA achieves $96.9$\% success with a mean completion length of $151$ steps. At the $1.5\times$ command, it achieves $96.8$\% over $111$ steps. These results correspond to a success-rate change of $-0.1$ percentage points and a realized rate of approximately $1.36\times$, calculated as $151/111$.

\subsection{Our Baseline and \method{}}

The vanilla resampling and \method{} results come from our own experiments using frozen pretrained checkpoints. Neither method requires additional policy training.

On LIBERO, we use the frozen $\pi_{0.5}$ checkpoint and evaluate all $40$ tasks across the four suites. TempoVLA is also evaluated on LIBERO with $\pi_{0.5}$, so using the same policy keeps the benchmark and the policy matched between the compared entries. The reference achieves $96.5$\% success. At the $2\times$ command, vanilla resampling achieves $87.3$\% success, a decrease of $9.2$ percentage points, with a realized rate of $1.55\times$. \method{} maintains the reference success rate of $96.5$\% while achieving a realized rate of $1.34\times$.

On Robomimic, we use state-input Diffusion Policy models \citep{diffusion_policy} pretrained through behavior cloning and released by the DPPO authors \citep{ren2025dppo}. We use these checkpoints without RL fine-tuning and keep their weights frozen throughout evaluation. We evaluate Lift, Can, and Square using $20$ denoising steps. The reference achieves $74.0$\% success. At the $2\times$ command, vanilla resampling reduces success by $8.0$ percentage points at a realized rate of $1.70\times$, while \method{} improves success by $2.67$ percentage points at $1.32\times$.

On CALVIN, we use the frozen $\pi_{0.5}$ checkpoint released by RLinf \citep{pi05,chen2025pi_rl, yu2026rlinf}, supervised fine-tuned on scenes A, B, and C and evaluated on scene D without additional policy training. Success requires completing all five instructions in a sequence, while realized rates are averaged equally over the $34$ task types. The reference achieves $63.4$\% success. At the $4\times$ command, vanilla resampling achieves $57.1$\% success, a decrease of $6.3$ percentage points, at a realized rate of $2.474\times$. \method{} improves success to $64.9$\%, an increase of $1.5$ percentage points, at $2.009\times$.

\section{Additional Results}
\label{app:additional_results}

\subsection{Additional Simulation Results}
\label{app:CALVIN_details}

\begin{table}[!htbp]
\centering
\setlength{\belowcaptionskip}{8pt}
\caption{\textbf{Completion steps under fixed-rate execution on 10 LIBERO-90 tasks with Flow Matching.} Each entry reports the mean number of completion steps over that condition's successful rollouts. Tasks are identified by their LIBERO-90 indices and described in Table~\ref{tab:sim_task_list}. Vanilla denotes vanilla resampling. Success rates and realized rates are reported in Table~\ref{tab:sim_fixed}.}
\label{tab:sim_fixed_steps}
\scriptsize
\renewcommand{\arraystretch}{0.9}
\setlength{\tabcolsep}{2pt}
\begin{tabularx}{\textwidth}{@{}>{\raggedright\arraybackslash}p{40pt} >{\raggedright\arraybackslash}p{22pt} !{\vrule width 0.45pt} >{\columncolor{SimLavender}}c !{\vrule width 0.45pt} Y Y >{\columncolor{RRCBlue}}Y !{\vrule width 0.35pt} Y Y >{\columncolor{RRCBlue}}Y !{\vrule width 0.35pt} Y >{\columncolor{RRCBlue}}Y !{\vrule width 0.35pt} Y >{\columncolor{RRCBlue}}Y@{}}
\toprule
\multirow{2}{*}{Type} & \multirow{2}{*}{Task} & \multicolumn{1}{!{\vrule width 0.45pt}c!{\vrule width 0.45pt}}{$1\times$} & \multicolumn{3}{c!{\vrule width 0.35pt}}{$1.5\times$} & \multicolumn{3}{c!{\vrule width 0.35pt}}{$2\times$} & \multicolumn{2}{c!{\vrule width 0.35pt}}{$0.5\times$} & \multicolumn{2}{c}{$0.3\times$} \\
\cmidrule(lr){4-6}\cmidrule(lr){7-9}\cmidrule(lr){10-11}\cmidrule(lr){12-13}
& & Ref. & \wocrstack & \vanillastack & \methodstack & \wocrstack & \vanillastack & \methodstack & \wocrstack & \methodstack & \wocrstack & \methodstack \\
\midrule
\multirow{4}{40pt}{\raggedright Structure\\constrained} & 2 & $121$ & $100$ & $98$ & $105$ & $119$ & $93$ & $100$ & $242$ & $194$ & $440$ & $284$ \\
 & 32 & $204$ & $148$ & $160$ & $178$ & $127$ & $125$ & $152$ & $364$ & $327$ & $580$ & $489$ \\
 & 40 & $203$ & $158$ & $176$ & $161$ & $161$ & $149$ & $175$ & $360$ & $338$ & $584$ & $560$ \\
 & 86 & $157$ & $127$ & $116$ & $123$ & $109$ & $105$ & $110$ & $279$ & $255$ & $448$ & $409$ \\
\midrule
\multirow{3}{40pt}{\raggedright Bounded\\receptacle} & 18 & $188$ & $138$ & $136$ & $153$ & $125$ & $112$ & $140$ & $336$ & $328$ & $557$ & $497$ \\
 & 48 & $160$ & $127$ & $130$ & $136$ & $108$ & $108$ & $126$ & $328$ & $285$ & $659$ & $472$ \\
 & 60 & $113$ & $98$ & $94$ & $92$ & $94$ & $106$ & $87$ & $222$ & $191$ & $445$ & $321$ \\
\midrule
\multirow{3}{40pt}{\raggedright Open\\relational} & 34 & $113$ & $116$ & $112$ & $94$ & $107$ & $92$ & $85$ & $189$ & $173$ & $338$ & $262$ \\
 & 65 & $173$ & $150$ & $143$ & $171$ & $105$ & $108$ & $164$ & $274$ & $231$ & $521$ & $376$ \\
 & 69 & $113$ & $110$ & $105$ & $99$ & $106$ & $95$ & $108$ & $221$ & $187$ & $462$ & $338$ \\
\midrule
\multicolumn{2}{@{}l}{Mean} & $154$ & $127$ & $127$ & $131$ & $116$ & $109$ & $125$ & $282$ & $251$ & $503$ & $401$ \\
\bottomrule
\end{tabularx}
\end{table}

Table~\ref{tab:sim_fixed_steps} reports the mean number of steps to completion across 10 LIBERO-90 tasks at each commanded rate, with each value averaged only over rollouts that succeed under the corresponding condition.

Tables~\ref{tab:robomimic_fixed} and \ref{tab:calvin_fixed} extend the fixed-stride evaluation to Robomimic and CALVIN, showing how success and realized execution rate change across acceleration and deceleration commands.

On Robomimic, \method{} achieves higher success than both baselines at every acceleration command. At $4\times$, it retains $55.3$\% success, compared with $25.3$\% for vanilla resampling and $10.0$\% for w/o CR. Increasing the command from $2\times$ to $4\times$ raises its realized rate from $1.315\times$ to $1.473\times$. Thus, adaptive rescaling mitigates the degradation caused by aggressive acceleration but does not eliminate it. Under deceleration, \method{} also consistently outperforms w/o CR. At $0.3\times$, it achieves $76.0$\% success at a realized rate of $0.355\times$, compared with $62.0$\% at $0.274\times$ for w/o CR.

\begin{table}[!t]
\centering
\setlength{\belowcaptionskip}{4pt}
\caption{\textbf{Fixed-stride execution on Robomimic with frozen Diffusion Policies.} Success is aggregated over Lift, Can, and Square using the original behavior-cloning checkpoints, with $50$ trials per task and condition. Rate is the equal-task mean of reference/method step ratios on jointly successful trials. Bold marks the highest success rate under each command.}
\label{tab:robomimic_fixed}
\scriptsize
\renewcommand{\arraystretch}{0.9}
\setlength{\tabcolsep}{2pt}
\begin{tabularx}{\textwidth}{@{}p{73pt} !{\vrule width 0.45pt} >{\columncolor{SimLavender}}Y !{\vrule width 0.45pt} Y Y >{\columncolor{RRCBlue}}Y !{\vrule width 0.35pt} Y Y >{\columncolor{RRCBlue}}Y !{\vrule width 0.35pt} Y Y >{\columncolor{RRCBlue}}Y@{}}
\toprule
\rowcolor{GroupBand}\multicolumn{11}{c}{\bfseries (a) Robomimic: speedup} \\
\midrule
 & \multicolumn{1}{c!{\vrule width 0.45pt}}{$1\times$} & \multicolumn{3}{c!{\vrule width 0.35pt}}{$2\times$} & \multicolumn{3}{c!{\vrule width 0.35pt}}{$3\times$} & \multicolumn{3}{c}{$4\times$} \\
\cmidrule(lr){3-5}\cmidrule(lr){6-8}\cmidrule(lr){9-11}
Statistic & Ref. & \wocrstack & \vanillastack & \methodstack & \wocrstack & \vanillastack & \methodstack & \wocrstack & \vanillastack & \methodstack \\
\midrule
\rowcolor{GroupBand}\multicolumn{11}{c}{\bfseries Overall summary: three tasks} \\
\midrule
Success (\%) & $74.0$ & $64.0$ & $66.0$ & $\mathbf{76.7}$ & $32.0$ & $35.3$ & $\mathbf{60.7}$ & $10.0$ & $25.3$ & $\mathbf{55.3}$ \\
Rate ($\times$) & $1.000$ & $1.811$ & $1.702$ & $1.315$ & $1.755$ & $1.545$ & $1.428$ & $2.124$ & $2.041$ & $1.473$ \\
\bottomrule
\end{tabularx}

\vspace{3pt}

\begin{tabularx}{\textwidth}{@{}p{73pt} !{\vrule width 0.45pt} >{\columncolor{SimLavender}}Y !{\vrule width 0.45pt} Y >{\columncolor{RRCBlue}}Y !{\vrule width 0.35pt} Y >{\columncolor{RRCBlue}}Y !{\vrule width 0.35pt} Y >{\columncolor{RRCBlue}}Y@{}}
\toprule
\rowcolor{GroupBand}\multicolumn{8}{c}{\bfseries (b) Robomimic: slowdown} \\
\midrule
 & \multicolumn{1}{c!{\vrule width 0.45pt}}{$1\times$} & \multicolumn{2}{c!{\vrule width 0.35pt}}{$0.5\times$} & \multicolumn{2}{c!{\vrule width 0.35pt}}{$0.3\times$} & \multicolumn{2}{c}{$0.2\times$} \\
\cmidrule(lr){3-4}\cmidrule(lr){5-6}\cmidrule(lr){7-8}
Statistic & Ref. & \wocrstack & \methodstack & \wocrstack & \methodstack & \wocrstack & \methodstack \\
\midrule
\rowcolor{GroupBand}\multicolumn{8}{c}{\bfseries Overall summary: three tasks} \\
\midrule
Success (\%) & $74.00$ & $66.67$ & $\mathbf{73.33}$ & $62.00$ & $\mathbf{76.00}$ & $58.67$ & $\mathbf{68.00}$ \\
Rate ($\times$) & $1.000$ & $0.451$ & $0.542$ & $0.274$ & $0.355$ & $0.167$ & $0.211$ \\
\bottomrule
\end{tabularx}
\par\vspace{2pt}
\begin{minipage}{\textwidth}
\end{minipage}
\end{table}

On CALVIN, \method{} preserves five-instruction success across the evaluated acceleration commands. As the command increases from $2\times$ to $4\times$, its realized rate rises from $1.388\times$ to $2.009\times$, while success remains between $63.4$\% and $64.9$\%, compared with the $63.4$\% reference. At $4\times$, vanilla resampling and w/o CR reach higher realized rates of $2.474\times$ and $2.276\times$, but their success rates fall to $57.1$\% and $54.5$\%, respectively. Under deceleration, \method{} realizes rates from $0.244\times$ to $0.569\times$, with success remaining within $1.3$ percentage points of the $63.4$\% reference. 

Table~\ref{tab:success_preserving_range} summarizes the success-preserving execution-rate endpoints for each policy. We treat each task-command pair as a setting and consider it success-preserving if FreeSpeed's empirical success rate matches or exceeds that of the $1\times$ reference on the same task. For each policy, the table reports the minimum and maximum realized rates among these settings, together with the corresponding task, command, and success rates of the reference and FreeSpeed. The $0.22\times$--$2.53\times$ range reported in the abstract spans the overall minimum and maximum across these settings. Both endpoints come from Fast-WAM, with observed success rates equal to their respective references.

Each panel of Figure~\ref{fig:teaser}b presents results for one evaluation subset: the ten LIBERO-90 tasks in Table~\ref{tab:sim_fixed} for Flow Matching; LIBERO-Object at $3\times$ and LIBERO-Long at $0.3\times$ for $\pi_{0.5}$; and LIBERO-Spatial at $2\times$ and LIBERO-Goal at $0.5\times$ for Fast-WAM.

\begin{table}[!t]
\centering
\setlength{\belowcaptionskip}{4pt}
\caption{\textbf{Fixed-stride execution on CALVIN with frozen $\pi_{0.5}$ (ABC$\rightarrow$D).} Each stride setting is evaluated over $1000$ trials with randomly sampled seeds. Success requires completing all five instructions in a sequence. Realized rates are computed from paired reference/method step ratios, with equal weighting across the $34$ task types. Bold marks the highest success rate under each command.}
\label{tab:calvin_fixed}
\scriptsize
\renewcommand{\arraystretch}{0.9}
\setlength{\tabcolsep}{2pt}
\begin{tabularx}{\textwidth}{@{}p{73pt} !{\vrule width 0.45pt} >{\columncolor{SimLavender}}Y !{\vrule width 0.45pt} Y Y >{\columncolor{RRCBlue}}Y !{\vrule width 0.35pt} Y Y >{\columncolor{RRCBlue}}Y !{\vrule width 0.35pt} Y Y >{\columncolor{RRCBlue}}Y@{}}
\toprule
\rowcolor{GroupBand}\multicolumn{11}{c}{\bfseries (a) CALVIN: speedup} \\
\midrule
 & \multicolumn{1}{c!{\vrule width 0.45pt}}{$1\times$} & \multicolumn{3}{c!{\vrule width 0.35pt}}{$2\times$} & \multicolumn{3}{c!{\vrule width 0.35pt}}{$3\times$} & \multicolumn{3}{c}{$4\times$} \\
\cmidrule(lr){3-5}\cmidrule(lr){6-8}\cmidrule(lr){9-11}
Statistic & Ref. & \wocrstack & \vanillastack & \methodstack & \wocrstack & \vanillastack & \methodstack & \wocrstack & \vanillastack & \methodstack \\
\midrule
\rowcolor{GroupBand}\multicolumn{11}{c}{\bfseries Overall summary: five-instruction chains, 34 task types} \\
\midrule
Success (\%) & $63.4$ & $62.7$ & $62.6$ & $\mathbf{63.4}$ & $59.8$ & $55.0$ & $\mathbf{64.7}$ & $54.5$ & $57.1$ & $\mathbf{64.9}$ \\
Rate ($\times$) & $1.000$ & $1.712$ & $1.668$ & $1.388$ & $2.038$ & $2.076$ & $1.778$ & $2.276$ & $2.474$ & $2.009$ \\
\bottomrule
\end{tabularx}

\vspace{3pt}

\begin{tabularx}{\textwidth}{@{}p{73pt} !{\vrule width 0.45pt} >{\columncolor{SimLavender}}Y !{\vrule width 0.45pt} Y >{\columncolor{RRCBlue}}Y !{\vrule width 0.35pt} Y >{\columncolor{RRCBlue}}Y !{\vrule width 0.35pt} Y >{\columncolor{RRCBlue}}Y@{}}
\toprule
\rowcolor{GroupBand}\multicolumn{8}{c}{\bfseries (b) CALVIN: slowdown} \\
\midrule
 & \multicolumn{1}{c!{\vrule width 0.45pt}}{$1\times$} & \multicolumn{2}{c!{\vrule width 0.35pt}}{$0.5\times$} & \multicolumn{2}{c!{\vrule width 0.35pt}}{$0.3\times$} & \multicolumn{2}{c}{$0.2\times$} \\
\cmidrule(lr){3-4}\cmidrule(lr){5-6}\cmidrule(lr){7-8}
Statistic & Ref. & \wocrstack & \methodstack & \wocrstack & \methodstack & \wocrstack & \methodstack \\
\midrule
\rowcolor{GroupBand}\multicolumn{8}{c}{\bfseries Overall summary: five-instruction chains, 34 task types} \\
\midrule
Success (\%) & $63.4$ & $62.5$ & $\mathbf{63.0}$ & $60.7$ & $\mathbf{62.0}$ & $62.5$ & $\mathbf{63.0}$ \\
Rate ($\times$) & $1.000$ & $0.509$ & $0.569$ & $0.347$ & $0.387$ & $0.208$ & $0.244$ \\
\bottomrule
\end{tabularx}
\par\vspace{2pt}
\begin{minipage}{\textwidth}
\end{minipage}
\end{table}
\FloatBarrier

\begin{table}[!htbp]
\centering
\setlength{\belowcaptionskip}{4pt}
\caption{\textbf{Endpoints of the per-task success-preserving rate range.} For each policy, we report the lowest and highest realized rates among task-command settings where FreeSpeed's observed success rate matches or exceeds the same task's $1\times$ reference. Success rates \(\%\) are based on $50$ rollouts per task and condition. Bold indicates the overall endpoints, $0.22\times$ and $2.53\times$.}
\label{tab:success_preserving_range}
\scriptsize
\renewcommand{\arraystretch}{1.15}
\setlength{\tabcolsep}{4pt}
\begin{tabular}{@{}lll cccc@{}}
\toprule
Endpoint & Policy & Task & Command & Realized rate & Ref. success (\%) & \method{} success (\%) \\
\midrule
\multirow{3}{*}{Minimum} & Flow Matching & LIBERO-90, task 48 & $0.3\times$ & $0.35\times$ & $92$ & $100$ \\
 & $\pi_{0.5}$ & LIBERO-10, task 6 & $0.2\times$ & $0.22\times$ & $92$ & $92$ \\
 & Fast-WAM & LIBERO-Spatial, task 6 & $0.2\times$ & $\mathbf{0.22\times}$ & $100$ & $100$ \\
\midrule
\multirow{3}{*}{Maximum} & Flow Matching & LIBERO-90, task 18 & $2\times$ & $1.37\times$ & $98$ & $98$ \\
 & $\pi_{0.5}$ & LIBERO-Goal, task 5 & $3\times$ & $2.17\times$ & $100$ & $100$ \\
 & Fast-WAM & LIBERO-Goal, task 0 & $4\times$ & $\mathbf{2.53\times}$ & $100$ & $100$ \\
\bottomrule
\end{tabular}
\end{table}
\FloatBarrier

\subsection{Constant-Factor Rescaling}
\label{app:constant_factor}

We conduct this sweep using frozen Flow Matching policies on LIBERO-90 tasks 34, 48, and 86, listed in Table~\ref{tab:sim_task_list}. We evaluate constant-factor rescaling at the $2\times$ command, with $50$ rollouts per task and condition. The rescaling factor $f$ remains constant throughout each rollout and is swept from $1.0$ to $2.0$ in increments of $0.1$. The setting $f=2$ coincides with w/o CR. All settings use the same resampling stride and number of actions executed per replan.

Across these three tasks and within the evaluated range, no constant factor matches both the mean success rate and the mean realized rate of \method{}. At a comparable realized rate, constant-factor rescaling achieves lower success; at a comparable success rate, it achieves a lower realized rate. Thus, selecting a single factor for the entire rollout does not recover the performance of \method{}.

\begin{figure}[!htbp]
  \centering
  \includegraphics[width=0.92\textwidth]{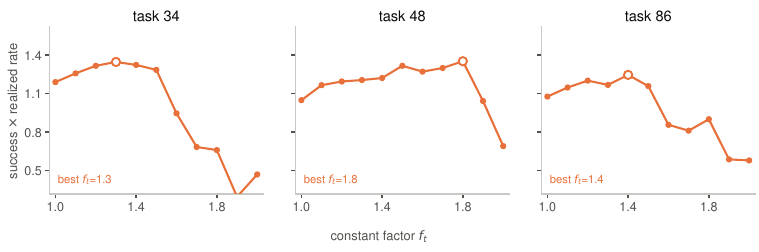}
  \caption{\textbf{Success-weighted realized rate for each task.} The joint score $J=S\hat{r}$ combines success rate $S\in[0,1]$ and realized rate $\hat{r}$. Each panel shows one task at the $2\times$ command, with $50$ rollouts per condition. The open marker identifies the highest-scoring constant factor within the sweep: $1.3$, $1.8$, and $1.4$ for tasks 34, 48, and 86, respectively.}
  \label{fig:constant_factor_tradeoff}
\end{figure}

In Figure~\ref{fig:constant_factor_tradeoff}, we further evaluate the joint score $J=S\hat{r}$, which combines success rate $S\in[0,1]$ and realized rate $\hat{r}$. The constant factors that maximize this score on tasks 34, 48, and 86 are $1.3$, $1.8$, and $1.4$, respectively. These task-dependent optima within the evaluated range indicate that obtaining the best success-speed trade-off through fixed-factor rescaling requires a separate factor sweep for each task. \method{} achieves comparable performance directly, without additional evaluation trials for per-task factor selection.

\begin{table}[!htbp]
\centering
\setlength{\belowcaptionskip}{4pt}
\caption{\textbf{Constant-factor versus adaptive rescaling.} Results on LIBERO-90 tasks 34, 48, and 86 at the $2\times$ command, with $50$ rollouts per task and condition. The factor $f$ remains constant throughout each rollout and is swept from $1.0$ to $2.0$ in increments of $0.1$. The setting $f=2$ corresponds to w/o CR. \method{} adapts the rescaling factor at each replan using Equation~\eqref{eq:factor_mapping}.}
\label{tab:constant_factor}
\scriptsize
\renewcommand{\arraystretch}{0.9}
\setlength{\tabcolsep}{3pt}
\begin{tabularx}{\textwidth}{@{}ll !{\vrule width 0.45pt} Y !{\vrule width 0.35pt} YYYYYYYYYYY !{\vrule width 0.35pt} Y@{}}
\toprule
& & \multicolumn{1}{c!{\vrule width 0.35pt}}{Ref.} & \multicolumn{11}{c!{\vrule width 0.35pt}}{\bfseries constant factor $f_t$} & \\
\cmidrule(lr){4-14}
Task & & $1\times$ & $1.0$ & $1.1$ & $1.2$ & $1.3$ & $1.4$ & $1.5$ & $1.6$ & $1.7$ & $1.8$ & $1.9$ & $2.0$ & \methodstack{} \\
\midrule
\multirow{2}{*}{34} & Succ. & $98$ & $100$ & $100$ & $100$ & $100$ & $96$ & $92$ & $84$ & $70$ & $62$ & $30$ & $40$ & $\mathbf{100}$ \\
& Rate & $1.000$ & $1.19$ & $1.26$ & $1.32$ & $1.35$ & $1.38$ & $1.40$ & $1.13$ & $0.98$ & $1.06$ & $0.99$ & $1.18$ & $1.36$ \\
\multirow{2}{*}{48} & Succ. & $92$ & $100$ & $100$ & $94$ & $92$ & $92$ & $96$ & $88$ & $88$ & $88$ & $72$ & $46$ & $\mathbf{98}$ \\
& Rate & $1.000$ & $1.05$ & $1.17$ & $1.27$ & $1.31$ & $1.33$ & $1.37$ & $1.45$ & $1.48$ & $1.54$ & $1.45$ & $1.50$ & $1.35$ \\
\multirow{2}{*}{86} & Succ. & $92$ & $92$ & $94$ & $92$ & $90$ & $88$ & $78$ & $56$ & $56$ & $54$ & $40$ & $40$ & $\mathbf{88}$ \\
& Rate & $1.000$ & $1.17$ & $1.22$ & $1.31$ & $1.30$ & $1.42$ & $1.49$ & $1.53$ & $1.45$ & $1.67$ & $1.47$ & $1.45$ & $1.43$ \\
\midrule
\rowcolor{RRCBlue} & Succ. & $94.0$ & $97.3$ & $98.0$ & $95.3$ & $94.0$ & $92.0$ & $88.7$ & $76.0$ & $71.3$ & $68.0$ & $47.3$ & $42.0$ & $\mathbf{95.3}$ \\
\rowcolor{RRCBlue}\multirow{-2}{*}{Mean} & Rate & $1.000$ & $1.14$ & $1.22$ & $1.30$ & $1.32$ & $1.37$ & $1.42$ & $1.37$ & $1.30$ & $1.42$ & $1.30$ & $1.38$ & $1.38$ \\
\bottomrule
\end{tabularx}
\end{table}

\begin{figure}[!htbp]
  \centering
  \includegraphics[width=0.54\textwidth]{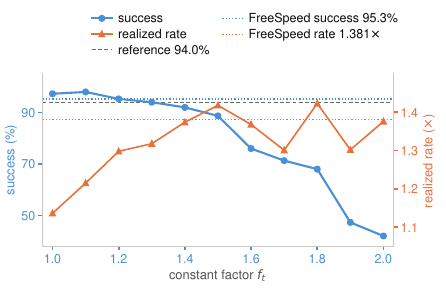}
  \caption{\textbf{Trade-off between success rate and realized rate under constant-factor rescaling.} Results are averaged equally across tasks 34, 48, and 86 at the $2\times$ command. The left and right axes show success rate and realized rate, respectively. Horizontal dashed lines indicate the results of \method{} and the $1\times$ reference.}
  \label{fig:constant_factor_curve_3task}
\end{figure}

\begin{figure}[!htbp]
  \centering
  \includegraphics[width=0.60\textwidth]{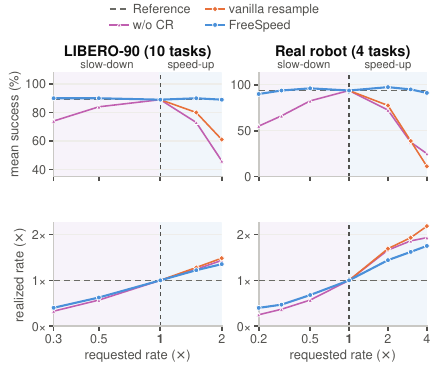}
  \caption{\textbf{Fixed-stride evaluations shown as curves.} Results from Tables~\ref{tab:sim_fixed} and~\ref{tab:real_fixed}, plotted against the requested execution rate. The upper row shows mean success rates, and the lower row shows realized execution rates. The curves illustrate how each method's success rate deviates from the $1\times$ reference as the requested rate changes.}
  \label{fig:real_curves}
\end{figure}

\subsection{Real-Robot Results in Detail}
\label{app:real_detail}

Table~\ref{tab:real_fixed_steps} provides the full real-robot results and reveals three additional findings that complement Section~\ref{sec:real_robot}.

First, deceleration can also substantially reduce task success. At the $0.5\times$, $0.3\times$, and $0.2\times$ commands, \method{} achieves mean success rates of $96.2$, $93.8$, and $90.0$\%, compared with $82.5$, $66.2$, and $55.0$\% for w/o CR. The gap widens as the commanded rate decreases, reaching $35$ percentage points at $0.2\times$. Slower execution therefore does not necessarily preserve success: repeating actions with proportionally reduced increments makes observations more densely spaced along the executed trajectory, which can shift the policy's inputs away from the training distribution.

Second, the baselines fail almost entirely on several tasks at the highest commands. At $4\times$, vanilla resampling records no successful trials on the peach, tennis-ball, or cube task. The w/o CR ablation also records no successes on the tennis-ball task and succeeds in only one of twenty trials on the cube task. In comparison, \method{} maintains success rates of $95$, $85$, and $90$\% on these three tasks, respectively. This degradation appears at a lower command for w/o CR, which already records no successes on the cube task at $3\times$.

Third, the higher realized rates of the baselines at $3\times$ and $4\times$ must be interpreted alongside their success rates. Realized rates are computed from successful rollouts, and tasks with no successful trials do not contribute to the reported average. At the $4\times$ command, vanilla resampling's $2.18\times$ rate therefore represents only the almond-pouring task, whereas \method{}'s $1.75\times$ averages over all four tasks. These values summarize different subsets of tasks and successful rollouts, so a higher reported rate alone does not establish better execution performance.

\begin{table}[!htbp]
\centering
\setlength{\belowcaptionskip}{8pt}
\caption{\textbf{Full fixed-rate results on four real-robot tasks with Flow Matching.} Entries report success rates in \%, mean completion steps over each condition's successful rollouts, and realized rates relative to the $1\times$ reference. A dash indicates that no trial succeeded. Success averages include all four tasks; completion-step and rate averages include only tasks with successful trials. Bold indicates the highest success rate under each command. Vanilla denotes vanilla resampling, which coincides with w/o CR under deceleration.}
\label{tab:real_fixed_steps}
\scriptsize
\renewcommand{\arraystretch}{0.9}
\setlength{\tabcolsep}{0.9pt}
\begin{tabularx}{\textwidth}{@{}p{34pt} !{\vrule width 0.45pt} >{\columncolor{SimLavender}}Y !{\vrule width 0.45pt} Y Y >{\columncolor{RRCBlue}}Y !{\vrule width 0.35pt} Y Y >{\columncolor{RRCBlue}}Y !{\vrule width 0.35pt} Y Y >{\columncolor{RRCBlue}}Y !{\vrule width 0.35pt} Y >{\columncolor{RRCBlue}}Y !{\vrule width 0.35pt} Y >{\columncolor{RRCBlue}}Y !{\vrule width 0.35pt} Y >{\columncolor{RRCBlue}}Y@{}}
\toprule
 & \multicolumn{1}{c!{\vrule width 0.45pt}}{$1\times$} & \multicolumn{3}{c!{\vrule width 0.35pt}}{$2\times$} & \multicolumn{3}{c!{\vrule width 0.35pt}}{$3\times$} & \multicolumn{3}{c!{\vrule width 0.35pt}}{$4\times$} & \multicolumn{2}{c!{\vrule width 0.35pt}}{$0.5\times$} & \multicolumn{2}{c!{\vrule width 0.35pt}}{$0.3\times$} & \multicolumn{2}{c}{$0.2\times$} \\
\cmidrule(lr){2-2}\cmidrule(lr){3-5}\cmidrule(lr){6-8}\cmidrule(lr){9-11}\cmidrule(lr){12-13}\cmidrule(lr){14-15}\cmidrule(lr){16-17}
 & Ref. & \wocrstack & \vanillastack & \methodstack & \wocrstack & \vanillastack & \methodstack & \wocrstack & \vanillastack & \methodstack & \wocrstack & \methodstack & \wocrstack & \methodstack & \wocrstack & \methodstack \\
\midrule
\rowcolor{GroupBand}\multicolumn{17}{c}{\bfseries Task 1: Peach} \\
\midrule
Succ. & $100$ & $70$ & $90$ & $\mathbf{100}$ & $50$ & $55$ & $\mathbf{100}$ & $45$ & $0$ & $\mathbf{95}$ & $70$ & $\mathbf{100}$ & $65$ & $\mathbf{100}$ & $20$ & $\mathbf{95}$ \\
Steps & $324$ & $219$ & $233$ & $234$ & $182$ & $211$ & $204$ & $206$ & -- & $202$ & $545$ & $433$ & $864$ & $665$ & $1252$ & $806$ \\
Rate & $1.00$ & $1.48$ & $1.39$ & $1.39$ & $1.77$ & $1.53$ & $1.59$ & $1.57$ & -- & $1.60$ & $0.59$ & $0.75$ & $0.38$ & $0.49$ & $0.26$ & $0.40$ \\
\midrule
\rowcolor{GroupBand}\multicolumn{17}{c}{\bfseries Task 2: Tennis} \\
\midrule
Succ. & $90$ & $90$ & $60$ & $\mathbf{95}$ & $45$ & $30$ & $\mathbf{90}$ & $0$ & $0$ & $\mathbf{85}$ & $80$ & $\mathbf{95}$ & $70$ & $\mathbf{90}$ & $65$ & $\mathbf{95}$ \\
Steps & $516$ & $285$ & $274$ & $367$ & $270$ & $230$ & $329$ & -- & -- & $292$ & $892$ & $748$ & $1191$ & $944$ & $1745$ & $1200$ \\
Rate & $1.00$ & $1.81$ & $1.89$ & $1.41$ & $1.91$ & $2.24$ & $1.57$ & -- & -- & $1.77$ & $0.58$ & $0.69$ & $0.43$ & $0.55$ & $0.30$ & $0.43$ \\
\midrule
\rowcolor{GroupBand}\multicolumn{17}{c}{\bfseries Task 3: Stack Cube} \\
\midrule
Succ. & $90$ & $60$ & $75$ & $\mathbf{95}$ & $0$ & $20$ & $\mathbf{90}$ & $5$ & $0$ & $\mathbf{90}$ & $\mathbf{95}$ & $90$ & $65$ & $\mathbf{95}$ & $\mathbf{75}$ & $\mathbf{75}$ \\
Steps & $482$ & $290$ & $261$ & $294$ & -- & $222$ & $271$ & $241$ & -- & $244$ & $831$ & $747$ & $1424$ & $1096$ & $2123$ & $1278$ \\
Rate & $1.00$ & $1.66$ & $1.84$ & $1.64$ & -- & $2.17$ & $1.78$ & $2.00$ & -- & $1.97$ & $0.58$ & $0.65$ & $0.34$ & $0.44$ & $0.23$ & $0.38$ \\
\midrule
\rowcolor{GroupBand}\multicolumn{17}{c}{\bfseries Task 4: Pour Almond} \\
\midrule
Succ. & $95$ & $70$ & $85$ & $\mathbf{100}$ & $55$ & $50$ & $\mathbf{100}$ & $50$ & $45$ & $\mathbf{95}$ & $85$ & $\mathbf{100}$ & $65$ & $\mathbf{90}$ & $60$ & $\mathbf{95}$ \\
Steps & $876$ & $522$ & $527$ & $651$ & $459$ & $496$ & $574$ & $394$ & $402$ & $527$ & $1669$ & $1337$ & $2630$ & $2098$ & $3717$ & $2306$ \\
Rate & $1.00$ & $1.68$ & $1.66$ & $1.34$ & $1.91$ & $1.77$ & $1.53$ & $2.23$ & $2.18$ & $1.66$ & $0.53$ & $0.66$ & $0.33$ & $0.42$ & $0.24$ & $0.38$ \\
\midrule
\rowcolor{GroupBand}\multicolumn{17}{c}{\bfseries Mean of the four tasks} \\
\midrule
Succ. & $93.8$ & $72.5$ & $77.5$ & $\mathbf{97.5}$ & $37.5$ & $38.8$ & $\mathbf{95.0}$ & $25.0$ & $11.2$ & $\mathbf{91.2}$ & $82.5$ & $\mathbf{96.2}$ & $66.2$ & $\mathbf{93.8}$ & $55.0$ & $\mathbf{90.0}$ \\
Steps & $549$ & $329$ & $324$ & $386$ & $304$ & $290$ & $344$ & $280$ & $402$ & $316$ & $984$ & $816$ & $1527$ & $1201$ & $2209$ & $1397$ \\
Rate & $1.00$ & $1.66$ & $1.69$ & $1.44$ & $1.86$ & $1.93$ & $1.62$ & $1.93$ & $2.18$ & $1.75$ & $0.57$ & $0.68$ & $0.37$ & $0.47$ & $0.25$ & $0.40$ \\
\bottomrule
\end{tabularx}
\end{table}

\subsection{Out-of-Distribution Measurements in Full}
\label{app:ood_full}

\begin{figure}[!t]
  \centering
  \includegraphics[width=\textwidth]{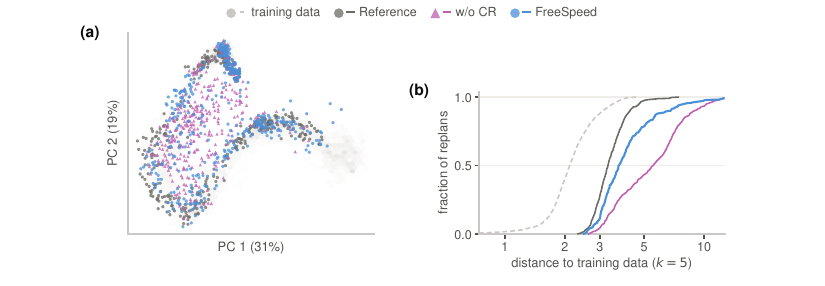}

  \caption{\textbf{Observation distribution shift under the $0.3\times$ command.} 
  Panels (a) and (b) repeat the principal-component projection and empirical distance CDF of Figure~\ref{fig:ood_grid} for deceleration. Distances are averaged over the $k=5$ nearest training vectors in the full conditioning space. Under deceleration, w/o CR coincides with vanilla resampling, so only three execution rules are shown.}
  \label{fig:ood_slow}

  \vspace{10pt}
  \includegraphics[width=\textwidth]{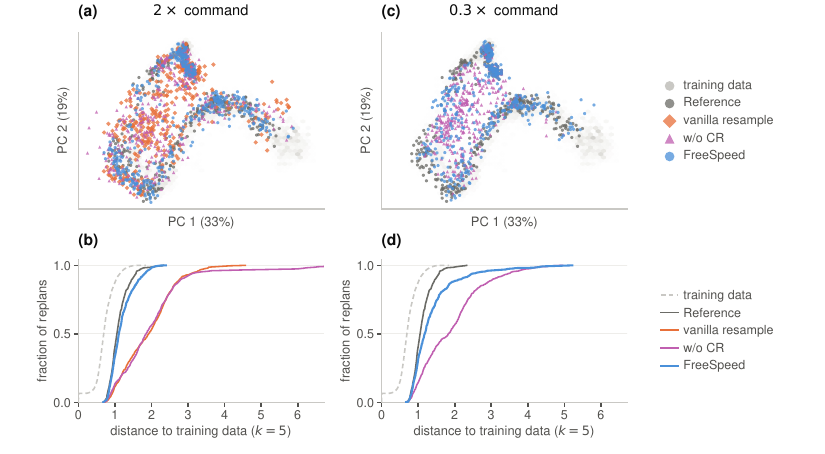}

  \caption{\textbf{Observation distribution shift measured using a single observation.} We repeat the analysis using only the final $384$ dimensions of the Flow Matching conditioning vector, which represent the current observation. Panels (a, b) show the $2\times$ command, and panels (c, d) show the $0.3\times$ command. Panels (a, c) show principal-component projections; panels (b, d) show empirical CDFs of mean distances to the $k=5$ nearest training vectors in this $384$-dimensional space.}

  \label{fig:ood_singleframe}
\end{figure}

This section defines the measurement procedure shared by Figures~\ref{fig:ood_grid}, \ref{fig:ood_slow}, and \ref{fig:ood_singleframe}, and presents two additional analyses.

\paragraph{Observation space.} All measurements use the policy's conditioning representation, consisting of the features produced by the observation encoder and passed to the policy.

The Flow Matching policy conditions on eight consecutive observations. Each observation contains six modalities: the language instruction, joint positions, gripper state, end-effector pose, a third-person image, and a wrist image. Each modality is encoded into a $64$-dimensional vector, yielding $384$ dimensions per observation. The eight representations are concatenated in temporal order to form a conditioning vector of $8\times384=3072$ dimensions. Its final $384$ dimensions correspond to the current observation. $\pi_{0.5}$ and Fast-WAM take a single observation as input, a setting examined further in Figure~\ref{fig:ood_singleframe}.

\paragraph{Training reference set and execution samples.} We construct the training reference set from the observations the policy receives in the human demonstrations. For each demonstration, we slide an eight-frame window along the trajectory with a stride of one frame. Each window contains the current frame and the seven preceding frames, matching the observation history used by the Flow Matching policy during execution. We encode each frame using the frozen policy's encoder and concatenate the eight representations into a $3072$-dimensional conditioning vector. The conditioning vectors collected across all demonstrations of a task form the \textbf{training reference set}. For Task 60, this set contains $6{,}176$ vectors, shown as the grey cloud in Figures~\ref{fig:ood_slow} and \ref{fig:ood_singleframe}.

Execution samples are collected from rollouts. For each execution method, we run 50 rollouts and record the conditioning vector passed to the policy at every replan, yielding one sample per replan. We determine a separate cutoff for each method using the median number of replans required to reach a selected task phase, such as the end of object placement. We then truncate that method's rollouts at this cutoff.

\paragraph{Principal-component projection.} The projection axes in Figure~\ref{fig:ood_grid}a are determined solely from the training data. We stack the conditioning vectors of all frames in the aforementioned training reference set as rows of an observation feature matrix. We then mean-center this matrix, compute its singular value decomposition \citep{pca}, and use the first two right singular vectors as the projection axes. Both the training data and all execution samples are projected onto these axes. For Task 60, the two components explain $31$\% and $19$\% of the variance, respectively.

Different methods produce different numbers of samples. For visualization, the panel displays the same number of points per method, equal to the smallest available sample count. All samples are retained for computing distances and cumulative curves. The projection illustrates the structure of the distributions, while quantitative comparisons are based on the distance measurements.

\paragraph{Distance to the training reference set.} For each execution sample, we compute its Euclidean distances to all training reference vectors in the full conditioning space and average the distances to its $k=5$ nearest neighbors \citep{pari2021surprising}. This gives one distance value per replan. Figure~\ref{fig:ood_grid}b plots the resulting distances for each method as an empirical cumulative distribution function (CDF). At each distance on the horizontal axis, the curve gives the fraction of replans whose distances are at or below that value. A curve further to the left therefore indicates that the method's conditioning vectors are closer to the training reference set. The reported medians correspond to the $50$\% quantile of these distances.

The grey dashed curve applies the same measurement to the training reference set itself, excluding self-matches. For each training vector, we retrieve its six nearest neighbors, exclude the vector itself, and average the distances to the remaining five.

Figures~\ref{fig:ood_grid}, \ref{fig:ood_slow}, and \ref{fig:ood_singleframe} display raw distances. In the main text and this section, we also report relative ratios, defined as each method's median distance divided by that of the $1\times$ reference on the same task. The reference therefore has a relative ratio of $1$ by construction. For Task 60, the reference median is $3.290$, and the median for the training set's self-comparison is $2.121$. At the $2\times$ command, w/o CR, vanilla resampling, and \method{} yield median distances of $5.532$, $5.886$, and $3.569$, corresponding to relative ratios of $1.682$, $1.789$, and $1.085$, respectively. At $0.3\times$, vanilla resampling and \method{} yield median distances of $5.532$ and $3.726$, corresponding to relative ratios of $1.680$ and $1.132$, respectively.

\paragraph{Inconsistency and subsequent distribution shift.} At each replan, we record the conditioning vector passed to the policy and the $I_{\cos}$ of the predicted chunk, computed using Equation~\eqref{eq:chunk_cosine_inconsistency}. Figure~\ref{fig:ood_grid}c pairs the $I_{\cos}$ at each replan with the distance to the training reference set at the next replan. This pairing relates chunk inconsistency to the deviation observed after chunk execution, which CI-based rescaling aims to limit. We compute quartile boundaries from the \(I_{\cos}\) distribution of the $1\times$ reference and apply the same boundaries to all execution methods. For each resulting bin, we report the median distance to the training reference set at the next replan.

\paragraph{Deceleration.} Figure~\ref{fig:ood_slow} repeats the first two panels of Figure~\ref{fig:ood_grid} at the $0.3\times$ command. Under deceleration, w/o CR uses $f_t=\rho_t$ and coincides with vanilla resampling, leaving two methods for comparison. The w/o CR ablation yields a median distance of $1.680$ times the reference median, while \method{} reduces this to $1.132$.

These results show that both acceleration and deceleration can shift the policy's conditioning inputs away from the training reference set \citep{romer2026failure}. Acceleration increases the spacing between consecutive observations along the executed trajectory, whereas deceleration reduces it. \method{} reduces the shift in both cases.

\paragraph{Single-observation analysis.} $\pi_{0.5}$ and Fast-WAM take a single observation as input. Figure~\ref{fig:ood_singleframe} examines whether the preceding results persist when the measurement is restricted to a single observation. We retain only the final $384$ dimensions of the Flow Matching conditioning vector, which encode the current observation's two images, proprioception, and language instruction, and repeat the same measurement.

At $2\times$, the relative distances of w/o CR, vanilla resampling, and \method{} change from $1.682$, $1.789$, and $1.085$ to $1.757$, $1.824$, and $1.050$, respectively. At $0.3\times$, the relative distances of w/o CR and \method{} change from $1.680$ and $1.132$ to $1.733$ and $1.119$, respectively. No value changes by more than $0.075$, and the gaps between the methods at $2\times$ are larger in the single-observation analysis. This analysis therefore reproduces the trend observed with the full conditioning vector, showing that the reduced deviation achieved by \method{} is also evident in the current observation alone.

\subsection{Keyframes from the Real-Robot Rollouts}
\label{app:real_keyframes}

\begin{figure}[p]
  \centering
  \includegraphics[width=0.88\textwidth,height=0.85\textheight,keepaspectratio]{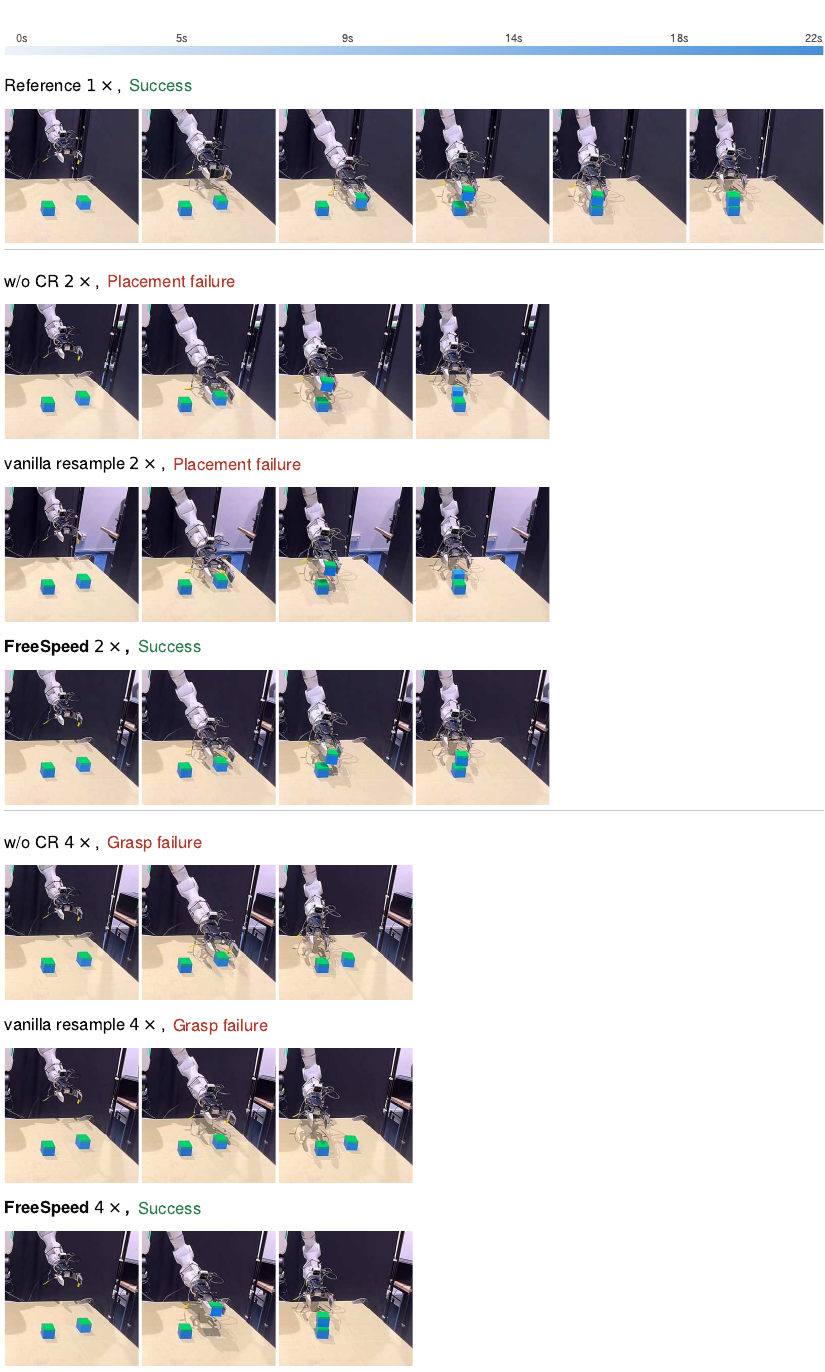}
  \caption{\textbf{Accelerated rollouts of the cube-stacking task.} One rollout per condition. The reference is sampled every $4.4$ seconds of wall-clock time, and each accelerated row is sampled at four points under the $2\times$ command and three under the $4\times$ one, spread over the episode.}
  \label{fig:keyframes_cube_accel}
\end{figure}

\begin{figure}[p]
  \centering
  \includegraphics[width=\textwidth,height=0.85\textheight,keepaspectratio]{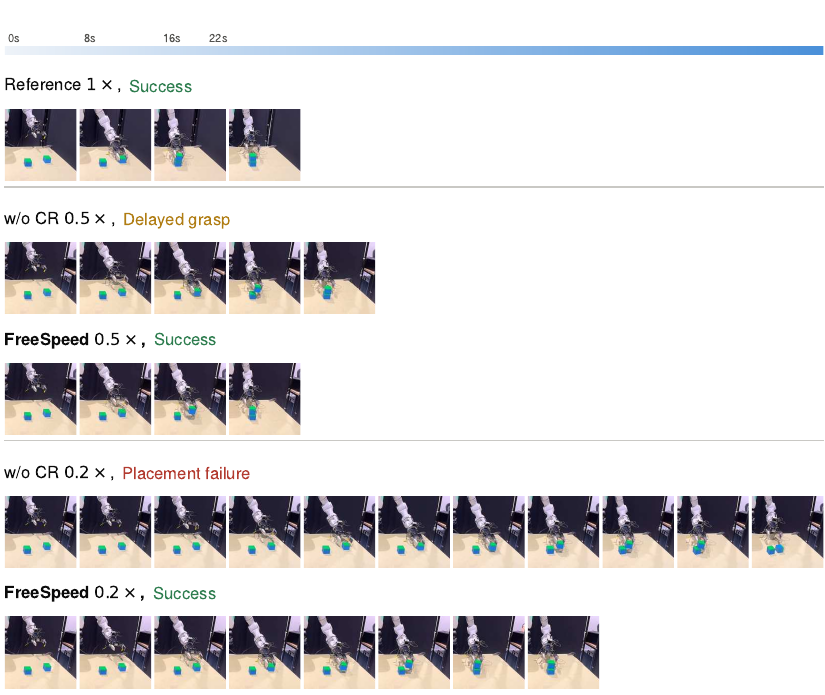}
  \caption{\textbf{Decelerated rollouts of the cube-stacking task.} One rollout per condition, sampled every $8.0$ seconds of wall-clock time.}
  \label{fig:keyframes_cube_decel}
\end{figure}

Figures~\ref{fig:keyframes_cube_accel} and~\ref{fig:keyframes_cube_decel} show one rollout of the cube-stacking task per condition as a strip of keyframes. In Figure~\ref{fig:keyframes_cube_decel}, frames are sampled at a fixed wall-clock interval, so strip length reflects episode duration. Each strip is annotated with the rollout outcome.

The failures are not all the same. At the $4\times$ command the two compared rules close the gripper without lifting the cube at all, so nothing later in the episode happens. At $2\times$ and at $0.2\times$ the cube is lifted but released beside the other one rather than on top of it. At $0.5\times$ the w/o CR ablation completes the stack, but only after the gripper hovers over the cube for several seconds. Every failure therefore happens at the grasp or at the release, the two moments where $I_{\cos}$ peaks in Figure~\ref{fig:demo_phase}a, and none happens while the arm is transporting the object.

These are single rollouts chosen to show the failure modes, not success rates. The success rates are in Table~\ref{tab:real_fixed} and Table~\ref{tab:real_fixed_steps}.

\begin{figure}[p]
  \centering
  \includegraphics[width=0.88\textwidth,height=0.85\textheight,keepaspectratio]{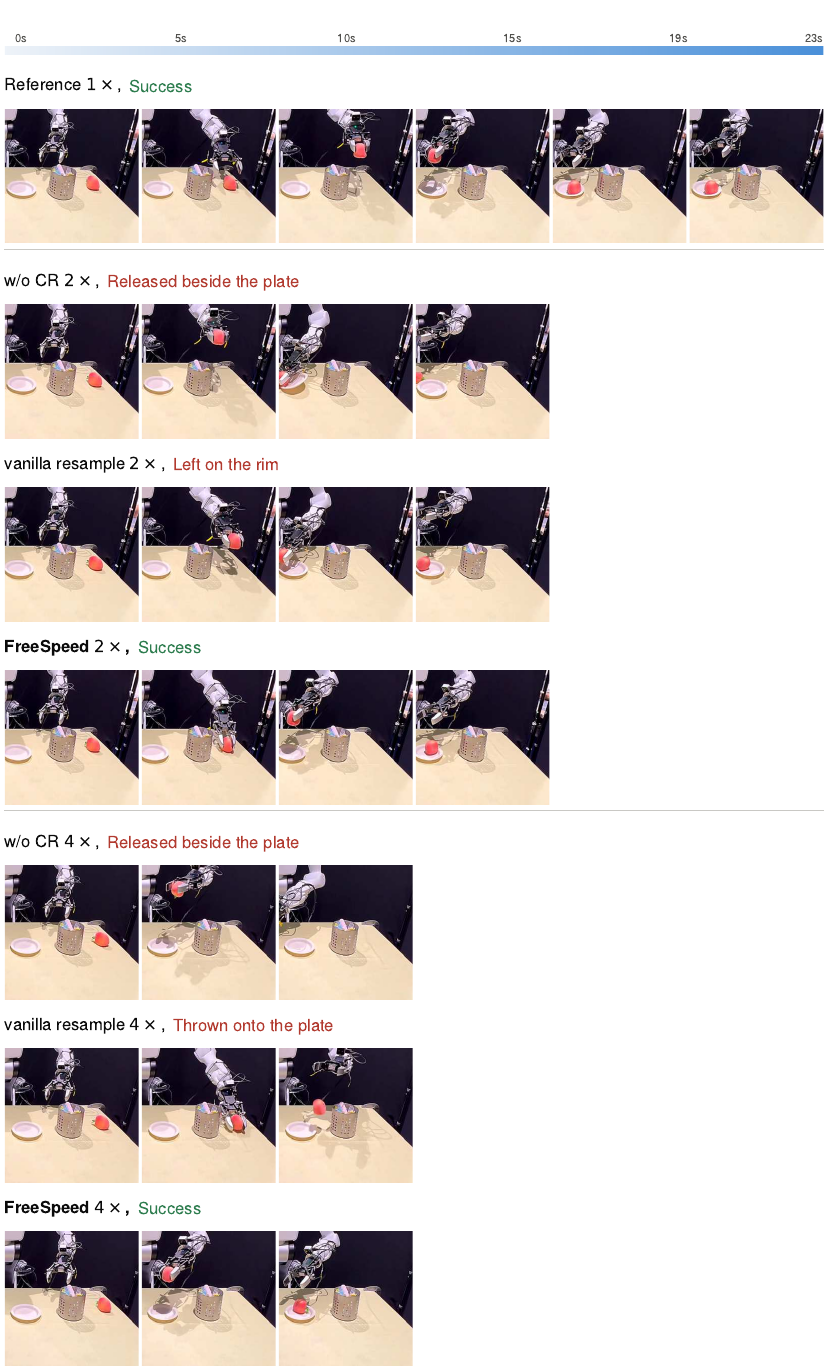}
  \caption{\textbf{Accelerated rollouts of the peach-placing task.} One rollout is shown per condition. The reference is sampled every $4.7$ seconds of wall-clock time. Each accelerated row shows four frames under the $2\times$ command and three under the $4\times$ command, sampled across the episode. At $4\times$, w/o CR releases the peach beside the plate, while vanilla resampling throws it; the final frame of the latter rollout shows the peach in the air.}
  \label{fig:keyframes_peach_accel}
\end{figure}

\begin{figure}[p]
  \centering
  \includegraphics[width=\textwidth,height=0.85\textheight,keepaspectratio]{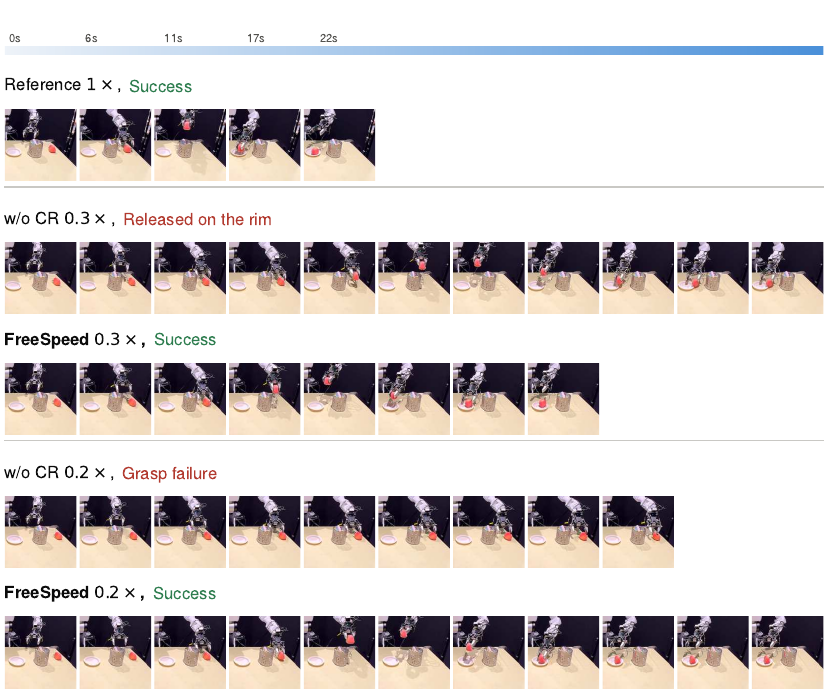}
  \caption{\textbf{Decelerated rollouts of the peach-placing task.} One rollout is shown per condition, sampled every $5.5$ seconds of wall-clock time. The displayed slowdown commands are $0.3\times$ and $0.2\times$. Under deceleration, vanilla resampling and w/o CR are equivalent, so only w/o CR is shown alongside \method{}.}
  \label{fig:keyframes_peach_decel}
\end{figure}

\begin{figure}[p]
  \centering
  \includegraphics[width=0.86\textwidth,height=0.85\textheight,keepaspectratio]{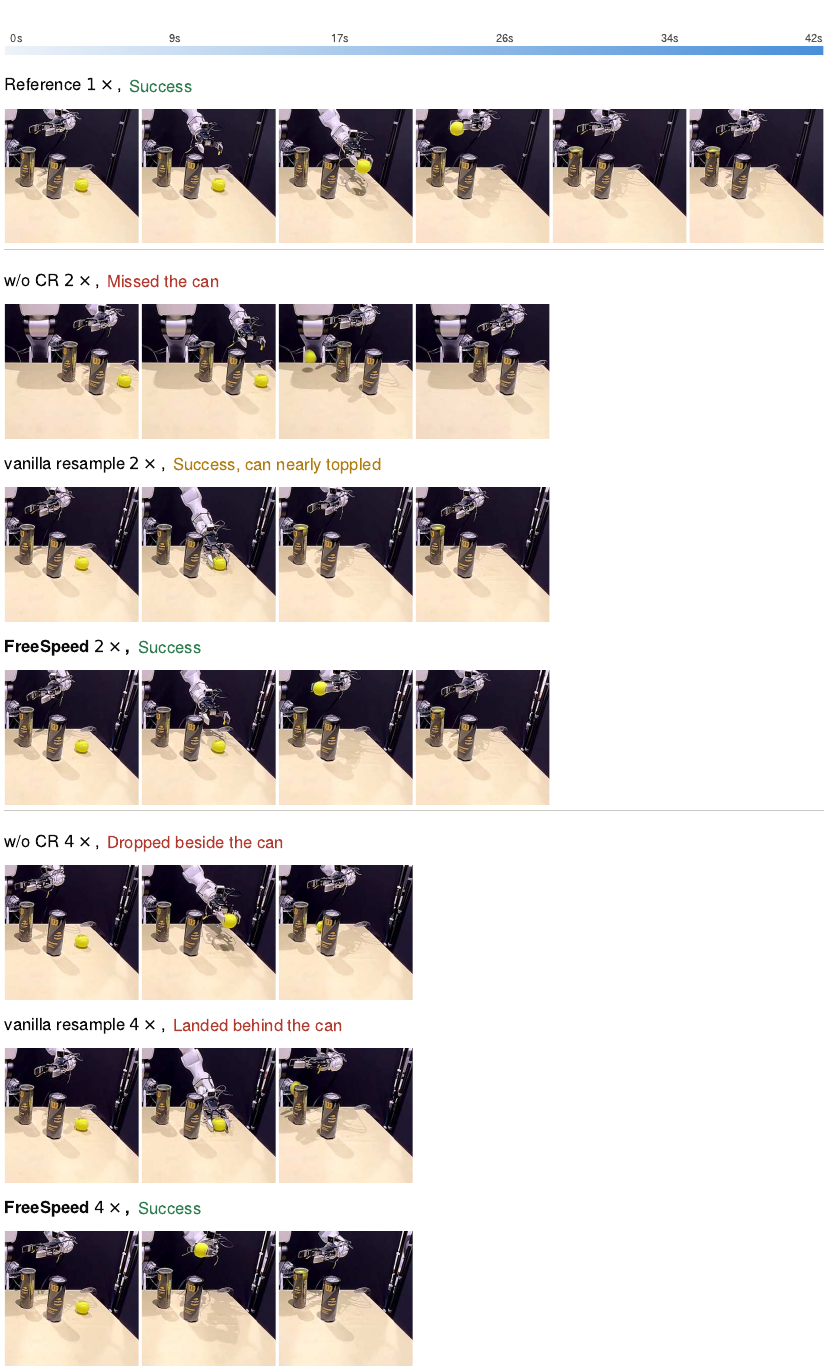}
  \caption{\textbf{Accelerated rollouts of the tennis-ball task.} One rollout is shown per condition. The reference is sampled every $8.4$ seconds of wall-clock time. Each accelerated row shows four frames under the $2\times$ command and three under the $4\times$ command, sampled across the episode. At $2\times$, vanilla resampling completes the task but nearly topples the can. At $4\times$, w/o CR drops the ball beside the can, while vanilla resampling leaves it behind the can, where it appears to overlap the opening from this camera angle.}
  \label{fig:keyframes_tennis_accel}
\end{figure}

\begin{figure}[p]
  \centering
  \includegraphics[width=\textwidth,height=0.85\textheight,keepaspectratio]{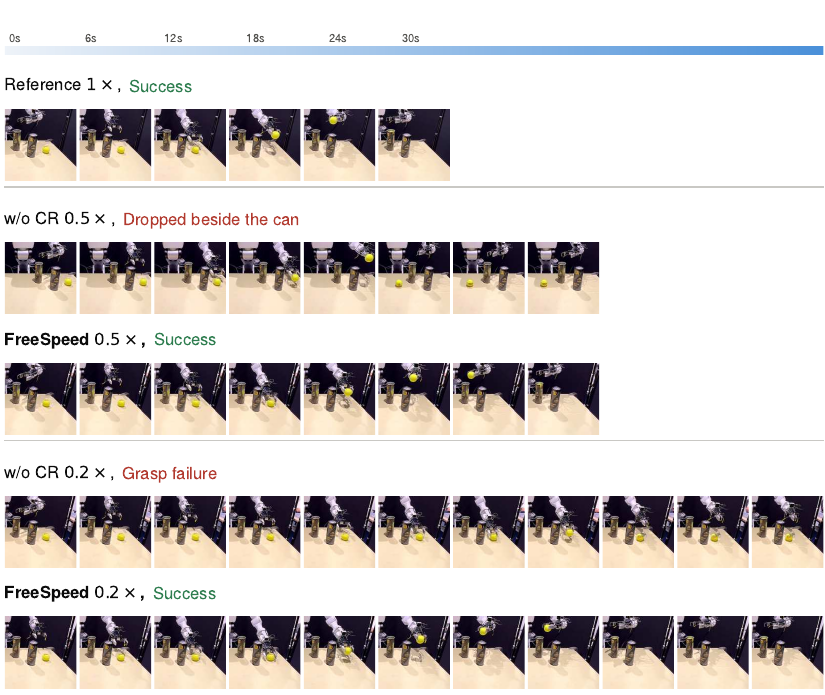}
  \caption{\textbf{Decelerated rollouts of the tennis-ball task.} One rollout is shown per condition, sampled every $5.9$ seconds of wall-clock time, with frames after the scene settles omitted. At $0.5\times$, vanilla resampling carries the ball over the can but releases it onto the table beside it.}
  \label{fig:keyframes_tennis_decel}
\end{figure}

\begin{figure}[p]
  \centering
  \includegraphics[width=0.88\textwidth,height=0.85\textheight,keepaspectratio]{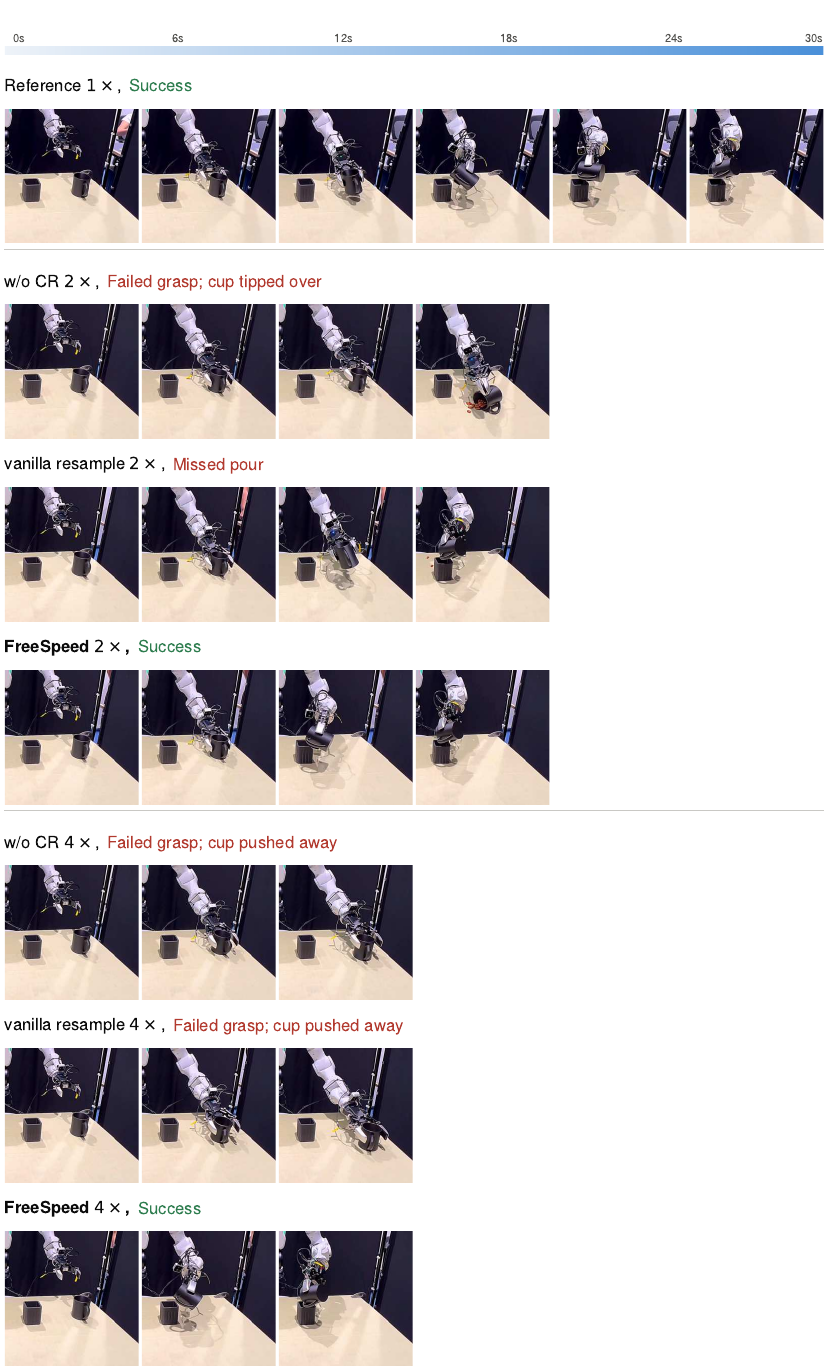}
  \caption{\textbf{Accelerated rollouts of the almond-pouring task.} One rollout per condition. The reference is sampled every $6.0$ seconds of wall-clock time, and each accelerated row is sampled at four points under the $2\times$ command and three under the $4\times$ one, spread over the episode.}
  \label{fig:keyframes_almond_accel}
\end{figure}

\begin{figure}[p]
  \centering
  \includegraphics[width=\textwidth,height=0.85\textheight,keepaspectratio]{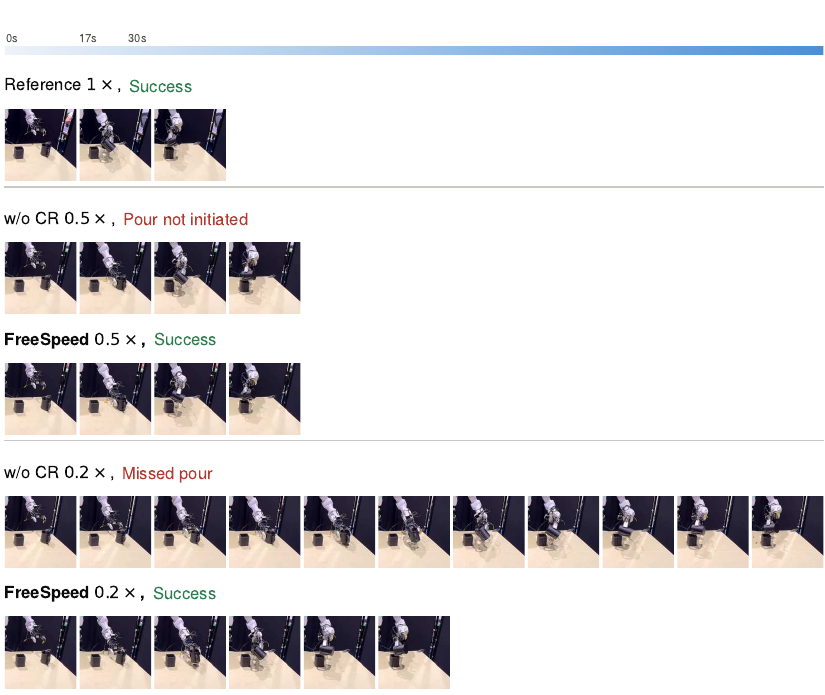}
  \caption{\textbf{Decelerated rollouts of the almond-pouring task.} One rollout per condition, sampled every $17.2$ seconds of wall-clock time.}
  \label{fig:keyframes_almond_decel}
\end{figure}

\clearpage

\section{Limitations}
\label{app:limitations}

\method{} cannot achieve every requested rate on every task. Across the four real-robot tasks, increasing the command from $3\times$ to $4\times$ raises the mean realized rate only from $1.62\times$ to $1.75\times$. The inconsistency-dependent rescaling amplifies action increments more aggressively when consecutive directions remain consistent and brings the factor toward $1$ as inconsistency increases. This limits step-length amplification around turns and during critical phases such as approach, grasping, and release, although temporal resampling still applies. The achievable acceleration therefore depends on how much of the task can be executed through directionally consistent motion, resulting in task-dependent limits on the realized speedup.
In addition, \method{} determines the rescaling factor solely from the kinematics of the predicted action chunk and does not use tactile or force feedback. Incorporating such physical-interaction signals \citep{shan2026learning} may further guide speed modulation during contact-rich phases.

\end{document}